\documentclass{bmvc2k}

\title{PERSIST: Persistent-State Discrimination for Shot Boundary Detection}

\addauthor{Tingyu Lin}{tylin@cvl.tuwien.ac.at}{1}
\addauthor{Christian Stippel}{}{1}
\addauthor{Armin Dadras}{}{1,2}
\addauthor{Jakob Zenzmaier}{}{3}
\addauthor{Florian Kleber}{}{1}
\addauthor{Wolfgang Aigner}{}{2}
\addauthor{Robert Sablatnig}{}{1}

\addinstitution{
 Computer Vision Lab\\
 TU Wien\\
 Vienna, Austria
}
\addinstitution{
 Institute of Creative\textbackslash Media/Technologies\\
 St.\ P\"olten University of Applied Sciences\\
 St.\ P\"olten, Austria
}
\addinstitution{
 Ludwig Boltzmann Institute for Digital History\\
 Ludwig Boltzmann Gesellschaft\\
 Vienna, Austria
}

\runninghead{Lin et al.}{PERSIST: Persistent-State SBD}

\def\eg{\emph{e.g}\bmvaOneDot}

\def\etal{\emph{et al}\bmvaOneDot}

\usepackage{booktabs}    
\usepackage{amsmath}     
\usepackage{amssymb}     
\usepackage{microtype}   
\usepackage{enumitem}    
\AtBeginDocument{%
  \setlength{\abovedisplayskip}{5pt plus 2pt minus 2pt}%
  \setlength{\belowdisplayskip}{5pt plus 2pt minus 2pt}%
  \setlength{\abovedisplayshortskip}{2pt plus 1pt}%
  \setlength{\belowdisplayshortskip}{2pt plus 1pt}}

\newcommand{\topone}[1]{\textbf{#1}}
\newcommand{\toptwo}[1]{\underline{#1}}

\begin{document}

\maketitle

\begin{abstract}
Shot boundary detection (SBD) is widely treated as the localisation of local visual discontinuities, yet many false positives such as hand-held shake, illumination flicker, motion blur, occlusion, and damaged archival material produce equally sharp local change without introducing a new shot. We reformulate SBD as boundary semantic discrimination: a frame is favoured as a boundary only when its local change evidence is accompanied by a persistent update of the video's latent temporal state, rather than a transient excursion that returns to the surrounding trend. This persistence test is operationalised with a continuous latent state from a FiLM-conditioned sinusoidal representation network and a structured discriminator that combines three semantic cues, local change, transient impulse, and return-to-trend, into a single interpretable per-frame signal over a dual-rate temporal backbone. The resulting framework, \mbox{\textsc{PERSIST}}, turns every decision into an inspectable one: the persistence criterion is trained into the classifier, its per-frame effect stays readable from the gate triple, and its learned latent state is measurably boundary-discriminative. On a $2{,}727$-video per-subtype diagnostic it removes $33$--$80\%$ of flash, text-overlay, and archival false positives relative to an identically trained cue detector, and at matched true-transition recall it roughly halves the official TransNetV2's pseudo-event false positives on that diagnostic and cuts its false positives on natural ClipShots footage by about a quarter, while preserving recall. It does so while reaching parity with the strongest public detector across modern online, broadcast, short-form, and historical-archive transfer evaluations, under markedly stricter training: it learns from ClipShots real transitions only, whereas the anchor draws on additional corpora whose transitions are $85\%$ synthetic. Code is available at \url{https://github.com/linty5/PERSIST}.
\end{abstract}

\section{Introduction}
\label{sec:intro}

\begin{figure*}[!t]
  \centering
  \includegraphics[width=0.9\linewidth]{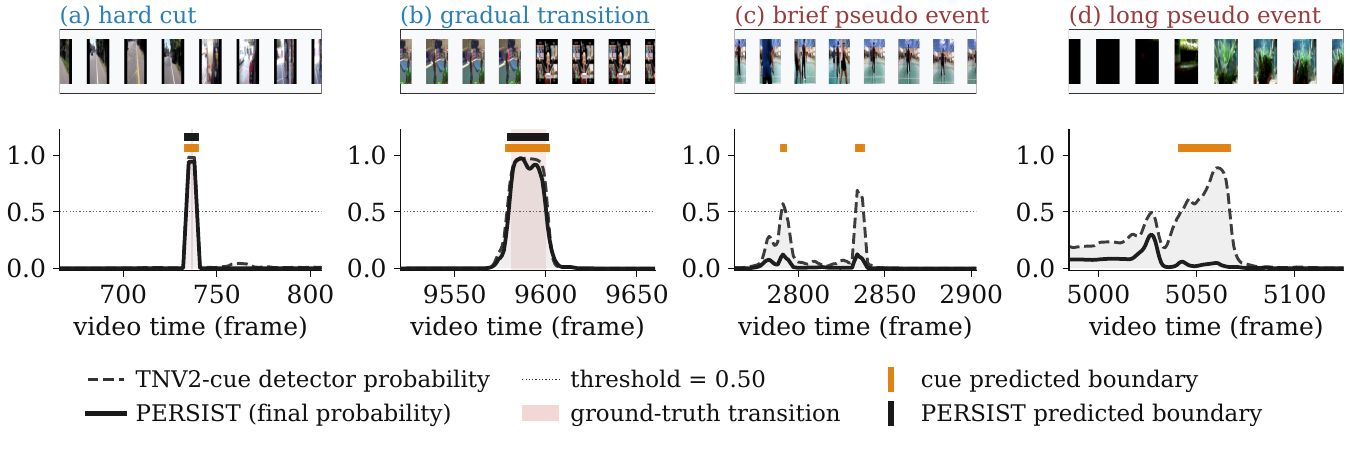}
  \caption{On four real ClipShots cases a TransNetV2-cue detector's per-frame probability (dashed) crosses threshold on all four including two pseudo events, whereas \textsc{PERSIST}'s probability (solid) fires only on the two genuine transitions.}
  \label{fig:teaser}
\end{figure*}

Shot boundary detection (SBD) locates the frames at which one continuously recorded shot ends and the next begins. As a basic form of temporal video segmentation, it is widely used as a front end for video indexing, retrieval, summarisation, editing, and higher-level video understanding~\cite{cotsaces2006review,kar2024sbdsurvey}. Modern benchmarks span online video~\cite{tang2018fast}, short-form video~\cite{zhu2023autoshot}, broadcast video~\cite{soucek2020transnetv2}, and historical archival film~\cite{helm2022historian}. Across these domains the hard cases are strikingly similar: hand-held shake, rapid object motion, motion blur, occlusion, illumination change, flicker, and damaged archival material can all create strong local discontinuities without introducing a new shot, as illustrated in Figure~\ref{fig:teaser}. Following terminology used in several prior systems, such non-transition events are referred to here as pseudo-boundaries.

Most deep SBD systems follow one effective recipe, encoding a short window, producing per-frame boundary scores, then thresholding and grouping~\cite{hassanien2017large,gygli2017ridiculously,soucek2019transnet,soucek2020transnetv2,zhu2023autoshot}, strengthened by hard-negative mining, frame-similarity cues, and harder benchmarks~\cite{hassanien2017large,soucek2020transnetv2,tang2018fast}. The central decision, however, is still driven by whether a local neighbourhood looks transition-like, and a local discontinuity, while necessary for a transition, is not sufficient: a hand-held shake peaks in frame-level distance signals, but the post-shake state matches the pre-shake state. We therefore reformulate SBD as boundary semantic discrimination: a frame is favoured as a boundary only if (i) it carries strong local change evidence, (ii) the change cannot be absorbed back into the surrounding visual trend, and (iii) it is registered persistently in the video's temporal state rather than as a short-lived impulse. This separates the few changes that move the video into a new persistent state from the many that do not.

The reformulation imposes three requirements, met by three co-designed components summarised in Figure~\ref{fig:method-overview}: a temporal representation in which persistence is well defined, a continuous latent state $\phi(t)$ from a sinusoidal representation network (SIREN)~\cite{sitzmann2020implicit} conditioned by feature-wise linear modulation (FiLM)~\cite{perez2018film}, whose analytic time derivatives shape it into a boundary-discriminative latent; a dual-rate backbone of TransNetV2-style cells~\cite{soucek2020transnetv2,feichtenhofer2019slowfast} supplying both short-range evidence and longer-range context; and a structured persistence discriminator that combines a change, a transient-impulse, and a return gate multiplicatively, so a candidate is down-weighted only when all three pseudo-boundary conditions hold jointly. Their product principally drives a training objective that teaches the classifier to separate persistent boundaries from transient pseudo-events, and at inference makes every decision readable through the gate triple. We call the framework \textsc{PERSIST}, a persistence test applied at every frame. The output stays a standard frame-level boundary score, so the framework is a drop-in replacement for any frame-level SBD detector and remains directly verifiable against existing SBD evaluation infrastructure.

Because the decision reads off a small set of per-frame functions, every suppression is attributable to a gate triple rather than hidden in an opaque score sum. This yields three results. The learned latent state is measurably boundary-discriminative, so the persistence test rests on a representation that genuinely encodes persistence. The mechanism then removes a large fraction of the flash, text-overlay, and archival false positives that motivate the paper while preserving true-transition recall. On standard benchmarks it matches the strongest public detector under markedly stricter training; and because aggregate SBD F1 has saturated and is dominated by ordinary transitions and background, the decisive question is the one it then answers, holding true-transition recall while removing the high-evidence pseudo-boundaries that drive deployment failures. The contribution is this reformulation and its inspectable mechanism, isolated from the backbone by the controlled comparison below.

\section{Related Work}
\label{sec:related}

This section reviews four lines of work that the proposed framework draws on or contrasts with. Scene boundary detection methods~\cite{rao2020lgss,chen2021shotcol,mun2022bassl,huang2020movienet} are cited but not directly compared, since scene segmentation groups multiple shots into higher-level narrative units rather than addressing the shot-level task here.

\paragraph{Classical and deep shot boundary detection.}
Classical SBD detects discontinuities with hand-crafted low-level cues, already noting that abrupt visual change is not necessarily a true transition: Zabih~\etal~\cite{zabih1995feature} use an edge-change ratio that accounts for camera and object motion, and Apostolidis and Mezaris~\cite{apostolidis2014fast} combine global and local colour descriptors. Deep methods replace hand-designed similarity with learned spatio-temporal representations: DeepSBD~\cite{hassanien2017large} classifies short segments with a 3D ConvNet trained on synthetic transitions and hard negatives; Gygli~\cite{gygli2017ridiculously} trains a fast fully convolutional detector on synthetic transitions; Tang~\etal~\cite{tang2018fast} introduce the ClipShots online-video benchmark and a structured cut/gradual model; TransNet and TransNetV2~\cite{soucek2019transnet,soucek2020transnetv2} establish strong efficient baselines from dilated 3D cells, synthetic rendering, frame-similarity cues, and multiple heads; and AutoShot~\cite{zhu2023autoshot} releases the SHOT short-video benchmark via architecture search.

\paragraph{Pseudo-boundary suppression and structured prediction.}
False positives from camera motion, illumination, and occlusion have been addressed at the data side, through hard-negative mining in DeepSBD~\cite{hassanien2017large}, or by additive cue fusion, as in the frame-similarity and colour-histogram heads of TransNetV2~\cite{soucek2020transnetv2} and the structured model of~\cite{tang2018fast}. We instead use a multiplicative joint gate over a learned continuous latent that is active only when change, transient-impulse, and return all hold, so every suppression decision is readable through a triple of gate values rather than hidden in an opaque score sum. Concurrent work redefines the prediction target, OmniShotCut~\cite{wang2026omnishotcut} predicts relational shot ranges and TransVLM~\cite{chen2026transvlm} continuous transition segments, each supervised on those targets, not frame-level boundaries, and reports on its own benchmarks, so it is not comparable to a frame-level detector under a shared evaluator; we deliberately retain the canonical frame-level output and position against the strongest publicly verifiable detector sharing it, TransNetV2. Outside SBD, structured boundary prediction recurs in temporal action localisation~\cite{lin2019bmn,lin2018bsn}, which builds candidates by aggregation rather than from a single peak.

\paragraph{Multi-rate and continuous temporal representations.}
SBD needs both short-range precision and longer-range context: a cut must be localised within a few frames, but rejecting shake, blur, flicker, and occlusion requires verifying whether the signal persists after the candidate. SlowFast networks~\cite{feichtenhofer2019slowfast} formalise a related separation in action recognition, while TransNetV2~\cite{soucek2020transnetv2} shows that efficient dilated 3D cells over low-resolution frames, in the lineage of C3D~\cite{tran2015c3d}, I3D~\cite{carreira2017quovadis}, and R(2+1)D~\cite{tran2018r2plus1d}, are well matched to SBD. Our backbone uses TransNetV2-style cells in both pathways but organises them into a dual-rate fast/slow structure with lateral fusion. For the latent state we adopt SIREN~\cite{sitzmann2020implicit}, whose sinusoidal parameterisation gives clean analytic first- and second-order time derivatives, with FiLM~\cite{perez2018film} conditioning from the slow pathway; this is what makes the persistence test computable. We use an explicit sinusoidal function rather than a state-space model such as S4~\cite{gu2022s4} or Mamba~\cite{gu2023mamba} precisely because the change, transient-impulse, and return cues are then directly readable from the function and its derivatives.

\paragraph{Benchmarks, domain transfer, and protocol reporting.}
Public SBD benchmarks span broadcast, online, and short-form video. TransNetV2~\cite{soucek2020transnetv2} reports results on ClipShots, BBC Planet Earth, and RAI, and AutoShot~\cite{zhu2023autoshot} reports transfer results on ClipShots and BBC alongside its SHOT benchmark. The main modern-video evaluation in this paper follows ClipShots, BBC Planet Earth, and AutoShot/SHOT; the choice of benchmark packages is detailed in the supplementary material.

Historical film provides a distinct transfer setting. Helm and Kampel~\cite{helm2019historicalfilms} highlight challenges such as damaged reels, scratches, and splices in digitised archival material, and Seidl~\etal~\cite{seidl2011gradual} study gradual-transition detection in historic film, where low-quality footage and long transition patterns make hand-crafted cues fragile. HISTORIAN~\cite{helm2022historian} provides digitised analogue film with expert cinematographic annotations. Because those annotations give shot boundaries rather than an explicit transition-type taxonomy, this paper reports HISTORIAN-GST12, a subset of HISTORIAN-derived videos with annotated gradual-transition spans, used to evaluate transfer where gradual-transition evidence is present.

\section{Method}
\label{sec:method}

\begin{figure*}[!tb]
  \centering
  \includegraphics[width=0.9\linewidth]{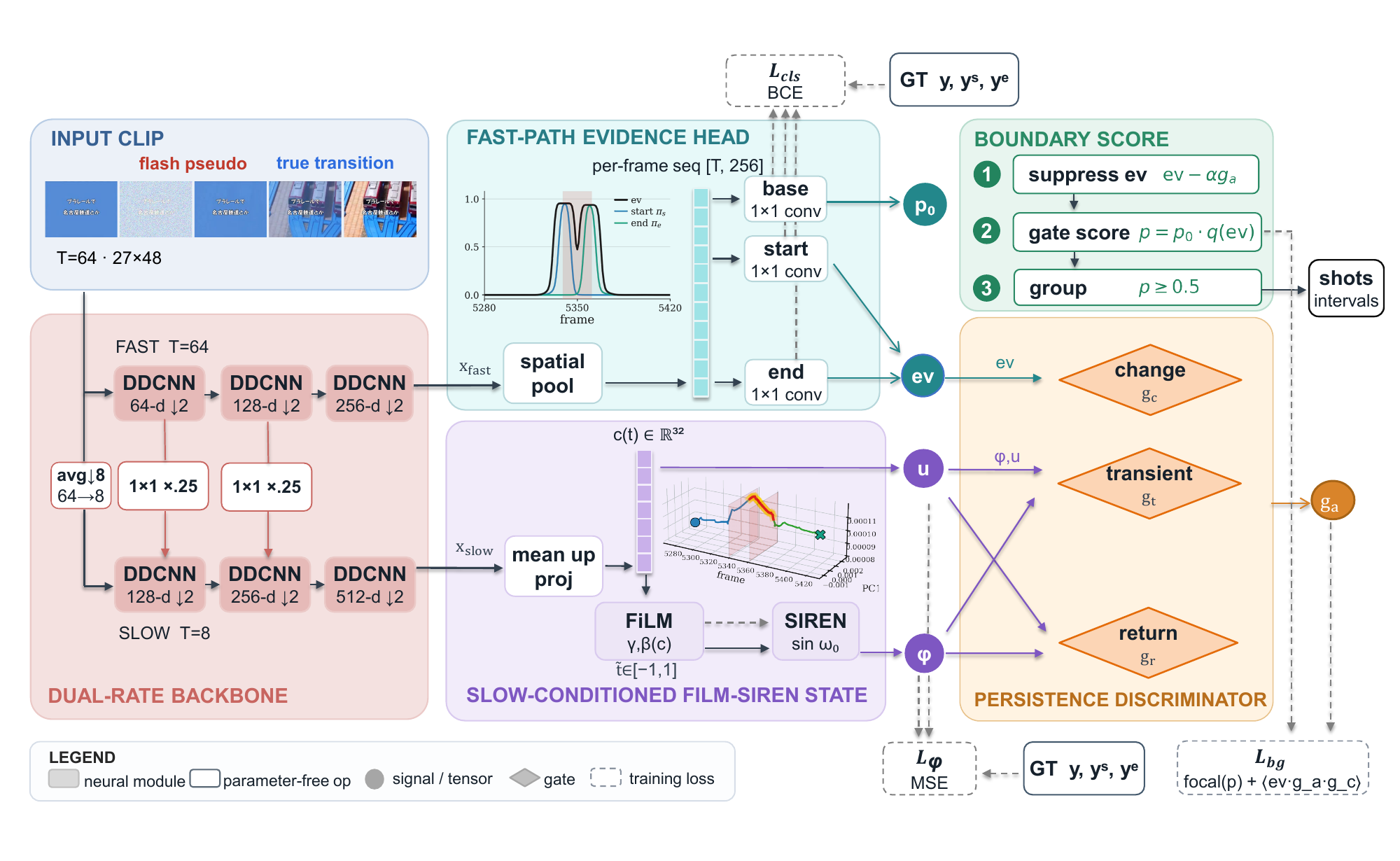}
  \caption{Architecture of \textsc{PERSIST} (left to right): a dual-rate TransNetV2-cell backbone feeds a fast-path evidence head ($\mathrm{ev}$, $p_0$) and a slow-conditioned FiLM-SIREN latent ($\phi$, teacher $u$), which a three-gate persistence discriminator ($g_c,g_t,g_r$) combines into a per-frame suppression $g_a$ that yields the final score $p$, with dashed nodes marking the training-only losses.}
  \label{fig:method-overview}
\end{figure*}

This section formalises boundary semantic discrimination, then specifies the three components of the proposed model: a dual-rate backbone, a FiLM-conditioned sinusoidal latent-state head, and a structured discriminator over the latent state. The complete pipeline is summarised in Figure~\ref{fig:method-overview}; subsequent subsections describe each component and the training objective in detail.

\subsection{Persistent-boundary criterion}
\label{sec:method-problem}

Let a video clip be a sequence of frames $V = (I_1, \ldots, I_T)$, $I_t \in \mathbb{R}^{H \times W \times 3}$. An SBD model is required to produce a per-frame boundary score $p(t) \in [0,1]$ from which a final list of shot intervals is obtained by thresholding and temporal grouping. In film-editing terms a shot is a maximal run of consecutive frames recorded without interruption, delimited by an in-point and an out-point; a shot boundary is the join between one out-point and the next in-point, produced either by in-camera starts and stops or by post-production editing, abrupt cuts and gradual transitions such as dissolves, fades, and wipes~\cite{reisz1968technique}. Abrupt cuts are instantaneous, whereas gradual transitions occupy a span of frames over which one shot gives way to the next, with both typically visible at once; a fade is the exception, spanning a single shot and the adjacent blank frames.

To make the distinction between true boundaries and pseudo-boundaries explicit, this paper postulates a continuous latent state $\phi: [1,T] \to \mathbb{R}^K$, of width $K{=}16$, that summarises the video at each instant. Because $\phi(t)$ is continuous, persistence is not defined by an infinitesimal jump at $t^\star$, but by finite-offset behaviour around the candidate. A frame index $t^\star$ is called a persistent boundary if, for a horizon set $\mathcal{H}$ of positive offsets fixed in advance and thresholds $0 < \varepsilon_{\text{d}} < \varepsilon_{\text{c}}$,
\begin{equation}
  \label{eq:persistent}
  \min_{d \in \mathcal{H}} \|\phi(t^\star + d) - \phi(t^\star - d)\| \;\ge\; \varepsilon_{\text{c}},
  \qquad
  \max_{d_1,d_2 \in \mathcal{H}^{+}} \|\phi(t^\star + d_1) - \phi(t^\star + d_2)\| \;\le\; \varepsilon_{\text{d}}.
\end{equation}
Here $\mathcal{H}$ contains positive offsets around the candidate and $\mathcal{H}^{+}$ contains offsets after the transition window, over which $d$, $d_1$, $d_2$ range. The first condition states that the latent state after $t^\star$ differs from the state before it at every offset in $\mathcal{H}$; the second states that, once the transition has elapsed, the post-boundary state has stabilised in the new regime. A pseudo-boundary is the negation of either condition: locally there is a sharp change, but the latent state either fails to update persistently at one of those offsets, or updates briefly and returns close to its pre-event value.

Equation~\eqref{eq:persistent} is conceptual; the model does not enforce it directly. Instead, it estimates $\phi(t)$ from the video clip and tests three operational proxies of the persistent-boundary condition that can be computed at every frame. The next three subsections describe how $\phi(t)$ is produced and how the proxies are evaluated.

\subsection{Dual-rate backbone}
\label{sec:method-backbone}

The backbone produces two feature streams from a clip downsampled to $27 \times 48$, $T = 64$ frames, built from TransNetV2-style temporal cells~\cite{soucek2020transnetv2}, the dilated 3D-convolution ``DDCNN'' blocks of Figure~\ref{fig:method-overview}, which use separable spatio-temporal kernels~\cite{tran2018r2plus1d}, batch normalisation~\cite{ioffe2015bn}, and residual connections~\cite{he2016resnet}. The fast pathway runs at native frame rate ($C_f{=}16$ base filters, three stages) and the slow pathway averages the input to stride $8$ ($C_s{=}32$, same depth), giving $\mathbf{F}_{\text{f}}$ at rate $T$ and $\mathbf{F}_{\text{s}}$ at rate $T/8$. For slow stage $l>0$, the previous fast-stage output is adaptively pooled in space and time, channel-projected by a $1\times1\times1$ convolution, and concatenated with the slow input:
\begin{equation}
  \label{eq:lateral}
  \mathbf{L}^{(l)} =
  \mathrm{Conv}_{1\times1\times1}\!\left(\mathrm{Pool}_{\text{st}}\!\left(\mathbf{F}_{\text{f}}^{(l-1)}\right)\right),
  \qquad
  \mathbf{F}_{\text{s}}^{(l)} =
  B_{\text{s}}^{(l)}\!\left([\mathbf{F}_{\text{s}}^{(l-1)}, \mathbf{L}^{(l)}]\right),
\end{equation}
where $\mathrm{Pool}_{\text{st}}$ is adaptive average pooling to the slow temporal rate and the slow spatial size, $[\cdot,\cdot]$ denotes channel concatenation, $B_{\text{s}}^{(l)}$ is the $l$-th slow TransNet-style stage, and the lateral channel width is $0.25$ times the corresponding fast-stage width. The final fast features carry short-range boundary evidence, while the final slow features carry trend context, enriched by pooled fast cues, that conditions the latent state in Section~\ref{sec:method-phi}. The fast features are spatially pooled to obtain a $T$-length vector sequence $\mathbf{F}_{\text{f}} \in \mathbb{R}^{T \times C_f^\star}$ that feeds three parallel $1{\times}1$ temporal convolutions producing per-frame start, end, and base logits. Writing $\pi_s(t), \pi_e(t)$ for the sigmoid of the start and end logits, the boundary evidence is
\begin{equation}
  \label{eq:ev}
  \mathrm{ev}(t) = \mathrm{maxpool}_{k}\!\left(\max\big(\pi_s(t), \pi_e(t)\big)\right),
\end{equation}
where $k{=}9$ is a temporal pooling kernel that lets nearby start/end candidates share evidence. We write $\ell_{\text{base}}(t)$ for the base logit and $p_0(t) = \sigma(\ell_{\text{base}}(t))$ for the corresponding base probability ($p_0$ in Figure~\ref{fig:method-overview}); both $\mathrm{ev}(t)$ and $p_0(t)$ are read by the persistence discriminator in Section~\ref{sec:method-trcs}.

\subsection{Latent state for persistence testing}
\label{sec:method-phi}

The slow pathway provides a per-frame conditioning signal $\mathbf{c}(t) \in \mathbb{R}^{D_c}$ with $D_c{=}32$. It is computed as a convex combination
$\mathbf{c}(t) = (1-\gamma)\,\bar{\mathbf{c}} + \gamma\,\tilde{\mathbf{c}}(t)$, where $\tilde{\mathbf{c}}(t)$ is the time-upsampled slow feature passed through a linear projection and a short temporal smoothing kernel, $\bar{\mathbf{c}}$ is its clip-level temporal average, and $\gamma\in(0,1)$ is a learned scalar. The FiLM-conditioned SIREN~\cite{sitzmann2020implicit,perez2018film} parameterises $\phi$ as a function of the normalised time index $\tilde{t} \in [-1,1]$ through $L{=}3$ hidden layers:
\begin{equation}
  \label{eq:siren}
  \mathbf{h}_l = \sin\!\Big( \omega_0 \big((1+\boldsymbol{\gamma}_l(\mathbf{c})) \odot (\mathbf{W}_l \mathbf{h}_{l-1} + \mathbf{b}_l) + \boldsymbol{\beta}_l(\mathbf{c})\big) \Big),
  \qquad
  \phi(\tilde t; \mathbf{c}) = \mathbf{W}_{L+1}\mathbf{h}_L + \mathbf{b}_{L+1} \in \mathbb{R}^{K},
\end{equation}
with $\mathbf{h}_0 = \tilde t$, output dimension $K{=}16$, and SIREN frequency $\omega_0{=}30$. The FiLM modulators $\boldsymbol{\gamma}_l(\mathbf{c}), \boldsymbol{\beta}_l(\mathbf{c})$ are produced by a per-layer linear hyper-network from $\mathbf{c}$. Because every non-linearity in~\eqref{eq:siren} is sinusoidal, the coordinate derivatives $\phi'(t)$ and $\phi''(t)$ are smooth and computed exactly through the chain rule while holding the conditioning vector fixed; they feed the training-time dyn loss in Section~\ref{sec:method-training}.

The conditioning signal is also passed through a separate learned linear projection
$\mathbf{u}(t) = \mathbf{W}_u\,\mathbf{c}(t) \in \mathbb{R}^{K}$
to give a teacher embedding in the same space as $\phi(t)$. $\mathbf{u}(t)$ has two roles: it is a distillation target for $\phi(t)$ during training, and it provides the observed-increment signal that the transient-impulse gate compares against (Section~\ref{sec:method-trcs}). An auxiliary scalar logit emitted from the SIREN body is used as a secondary classifier during training only.

\subsection{Persistence discriminator}
\label{sec:method-trcs}

The persistence discriminator turns Eq.~\eqref{eq:persistent} into three differentiable gates evaluated at every frame: a change gate that restricts suppression to frames carrying strong local evidence; a transient-impulse gate that fires when the observed step moves but the persistent latent does not, the signature of a non-absorbed pseudo-event; and a return gate that fires when the post-candidate latent has returned close to its pre-event configuration. Their product has, by design, two roles of unequal weight: its primary role is a background-suppression loss during training (Section~\ref{sec:method-training}) that teaches the base classifier to down-weight evidence-bearing frames flagged as pseudo-events, while at inference it deliberately contributes only a small evidence adjustment and a per-frame, per-cue attribution of the decision. The criterion is thus a training signal that shapes the classifier rather than a selective inference-time filter, a separation the gate-firing analysis in Section~\ref{sec:res-latent} then quantifies.

Let $\hat{\phi}(t) = \phi(t) / \|\phi(t)\|_2$ denote the normalised latent state and $\Delta_x f(t) = f(t+1) - f(t)$ a one-step forward difference operator. Each of the three cues is first reduced to a non-negative scalar signal $x_i(t)$ that quantifies one specific symptom of a pseudo-boundary; all three are then passed through the same soft-thresholding gate
\begin{equation}
  \label{eq:gate-form}
  g_i(t) = \sigma\!\left(\frac{x_i(t) - \theta_i}{\tau_i}\right),
  \qquad i \in \{\text{c},\,\text{t},\,\text{r}\},
\end{equation}
with per-cue threshold $\theta_i$ and temperature $\tau_i$. The three input signals are:
\begin{description}
  \item[\textbf{Change}, $x_{\text{c}}(t) = \mathrm{ev}(t)$.] Local boundary evidence from Eq.~\eqref{eq:ev}. A frame with $g_{\text{c}}(t)\!\approx\!0$ is not a candidate and is left untouched by the persistence discriminator.
  \item[\textbf{Transient-impulse}, $x_{\text{t}}(t) = I(t)$.] A true boundary peak should be matched by a comparable update of the persistent latent state, while a pseudo-boundary produces a peak that the latent state fails to absorb. The residual increment between the observation $\mathbf{u}(t)$ and the latent $\phi(t)$ is measured as:
    \begin{equation}
      \label{eq:innov}
      I(t) = \mathrm{ReLU}\!\big(\,\|\Delta_x \mathbf{u}(t)\|_2 \,-\, \|\Delta_x \phi(t)\|_2\,\big).
    \end{equation}
  $I(t)$ is large precisely when the observed step exceeds the latent step.
  \item[\textbf{Return}, $x_{\text{r}}(t) = S(t)$.] The first clause of Eq.~\eqref{eq:persistent} fails when, at one of the offsets considered, the latent state shortly after the candidate has returned close to the state shortly before it. This is captured through a multi-scale cosine similarity at offsets $\mathcal{D} = \{4, 8, 12\}$:
    \begin{equation}
      \label{eq:return}
      S(t) = \max_{d \in \mathcal{D}} \langle \hat{\phi}(t-d),\; \hat{\phi}(t+d) \rangle.
    \end{equation}
  The multi-scale max removes the need to tune a single per-dataset offset; $g_{\text{r}}(t)\!\approx\!1$ indicates the latent state has returned to a pre-event-like configuration.
\end{description}
The three gates are aggregated multiplicatively, so that the joint suppression signal is non-zero only if a candidate satisfies all three pseudo-boundary conditions jointly:
\begin{equation}
  \label{eq:gate-product}
  g_{\text{a}}(t) = g_{\text{c}}(t)\, g_{\text{t}}(t)\, g_{\text{r}}(t),
  \qquad
  \widetilde{\mathrm{ev}}(t) = \mathrm{clip}_{[0,1]}\!\big(\mathrm{ev}(t) - \alpha\, g_{\text{a}}(t)\big),
\end{equation}
where $\alpha$ is a fixed suppression strength. The suppressed evidence then re-scales the base probability through a learned boundary-factor blending parameter $\beta$:
\begin{equation}
  \label{eq:final-score}
  q(t) = (1-\beta) + \beta\,\widetilde{\mathrm{ev}}(t),
  \qquad
  p(t) = p_0(t)\, q(t).
\end{equation}
The final per-frame score stored for evaluation is $p(t)$; the training logit is recovered from $p(t)$ and post-shifted by a learned bias and temperature (see Section~\ref{sec:method-training}). The final list of intervals is obtained by thresholding $p(t)$ and applying the standard temporal grouping rule.

The product structure in~\eqref{eq:gate-product} is conservative by construction: a candidate has to satisfy the change, transient-impulse, and return conditions jointly to be down-weighted, and failing any one of the three leaves $\widetilde{\mathrm{ev}}$ close to $\mathrm{ev}$. Each suppression event can therefore be attributed to a triple $(g_{\text{c}}, g_{\text{t}}, g_{\text{r}})$, which we visualise per frame in Section~\ref{sec:res-latent}.

\subsection{Training objective and inference}
\label{sec:method-training}

The training loss decomposes into three groups by purpose,
\begin{equation}
  \label{eq:total-loss}
  \mathcal{L} = \mathcal{L}_{\text{cls}} + \mathcal{L}_{\phi} + \mathcal{L}_{\text{bg}},
\end{equation}
each a sum of standard terms whose full expressions and weights are given in the supplementary material. $\mathcal{L}_{\text{cls}}$ is a binary cross-entropy that supervises the main, base, SIREN, and start/end logits against the per-frame boundary label $y_t$, with a larger positive weight on the start/end heads. $\mathcal{L}_{\phi}$ shapes the latent state: it distils $\phi(t)$ towards the teacher $\mathbf{u}(t)$, penalises its temporal variation, and adds a dynamics term that makes the analytic energy $E(t){=}\|\phi'(t)\|_2{+}\mu\|\phi''(t)\|_2$ boundary-discriminative, where the SIREN chain-rule derivatives avoid the noise of frame-feature finite differences. $\mathcal{L}_{\text{bg}}$ suppresses background false positives with a focal penalty on $p$ and a term $\langle \mathrm{ev}\,\mathrm{sg}(g_a\,g_c)\rangle_{\text{bg}}$ ($\mathrm{sg}$: stop-gradient) that down-weights raw evidence the discriminator flags as a pseudo-event. At inference only $p_0(t)$, $\mathrm{ev}(t)$, the SIREN state, and the three gates are used, giving $\widetilde{\mathrm{ev}}(t)$ and then $p(t)$ via Eq.~\eqref{eq:final-score}; the auxiliary SIREN and dynamics logits are not evaluated, and the boundary list follows from thresholding $p(t)$ with standard temporal grouping.

\section{Experiments}
\label{sec:experiments}

The experiments establish three results in turn: where \textsc{PERSIST} stands against the literature and a matched baseline on standard benchmarks, and how much of any gap is the persistence mechanism rather than the backbone (Section~\ref{sec:res-bench}); which components matter, and in what sense (Section~\ref{sec:res-tier3}); and whether the mechanism suppresses pseudo-events and shapes a boundary-discriminative latent as the reformulation claims (Sections~\ref{sec:res-pertype} and~\ref{sec:res-latent}).

\subsection{Setup and evaluation protocol}
\label{sec:exp-setup}
\label{sec:exp-impl}
\label{sec:exp-tiers}

\paragraph{Data and protocol.}
We train on the ClipShots~\cite{tang2018fast} train split, using its real transitions only with no synthetic rendering or auxiliary corpus, and evaluate, with a single model and no per-target fine-tuning, on four test sets in decreasing in-domain similarity: ClipShots test (online video), BBC Planet Earth~\cite{soucek2020transnetv2} (broadcast), AutoShot/SHOT~\cite{zhu2023autoshot} (short-form), and HISTORIAN-GST12 (twelve HISTORIAN-derived archival videos~\cite{helm2022historian} with annotated gradual-transition spans); the gradual-transition proportion ranges from near $0\%$ on BBC to over $20\%$ on AutoShot. F1 is computed against the ground-truth intervals with the standard SBD matching rule ($\pm2$-frame tolerance), under two protocols: a fixed threshold of $0.50$ and an oracle-best threshold swept on the target split. Clips are resized to $27{\times}48$ and processed in $T{=}64$ sliding windows; the full transition-type taxonomy, per-set statistics, hyperparameters, and training details are in the supplementary material.

\paragraph{Comparison tiers.}
The experiments run in three tiers over one evaluation pipeline. \textbf{Tier~1} positions \textsc{PERSIST} against the official TransNetV2 public model~\cite{soucek2020transnetv2} and classical/DeepSBD baselines~\cite{zabih1995feature,apostolidis2014fast,hassanien2017large,tang2018fast}; here the training asymmetry runs against us, as TransNetV2 draws on additional corpora with $\sim$$85\%$ synthetic transitions while we use ClipShots real transitions only. \textbf{Tier~2} removes that asymmetry with a cue-only baseline sharing \textsc{PERSIST}'s backbone, optimiser, schedule, and seed, swapping only the persistence mechanism for the TransNetV2 descriptor-cue stack. \textbf{Tier~3} ablates one component at a time. Caveats on the excluded concurrent methods (Section~\ref{sec:related}) are in the supplementary material.

\paragraph{Checkpoint selection and seed disclosure.}
All numbers use one rule: the epoch with best validation F1 at the validation-swept threshold (``val-best''), reported under both protocols; the checkpoint sweep is on validation only, so checkpoint selection never sees the test split. The \textsc{PERSIST} row of Table~\ref{tab:main} is the three-seed mean (seeds $11$, $1$, $21$; per-seed standard deviation at most $0.0095$ fixed-$0.50$, $0.0064$ oracle-best, full table in the supplementary material), whereas the controlled comparisons of Tables~\ref{tab:controlled} and~\ref{tab:ablation} report the three-seed mean\,$\pm$\,population std over the same seeds, so that their rows differ only in the intended factor; per-seed values are in the supplementary material. A three-seed probability-average ensemble, reported separately in Table~\ref{tab:main} and kept out of the single-model comparison, lifts every dataset (supplementary material).

\subsection{Benchmark comparison and controlled attribution}
\label{sec:res-bench}

\paragraph{Literature positioning (Tier~1).}
Table~\ref{tab:main} reports the literature-level comparison on ClipShots, BBC Planet Earth, AutoShot/SHOT, and HISTORIAN-GST12. The official TransNetV2 (TNV2) PyTorch model is the strongest public anchor, alongside three classical baselines (pixel difference, HSV/$\chi^2$ histogram distance, edge change ratio) and the released DeepSBD AlexNet and ResNet-18 checkpoints of the Hassanien~\etal\ and Tang~\etal\ designs, run through our evaluator. All rows share the same matching tolerance and the same fixed-$0.50$ and oracle-best protocols, run through the identical matching pipeline; training data and frameworks vary across rows, as noted in Section~\ref{sec:exp-tiers}. The AutoShot released model is excluded for training-set overlap with the evaluation suite (supplementary material).

A single ClipShots-trained \textsc{PERSIST} matches the official TransNetV2 across all four datasets: as a three-seed mean it equals or exceeds the anchor on five of the eight cells, is within $0.003$ on BBC oracle-best, and trails beyond that only on AutoShot/SHOT ($-0.0115$ fixed-$0.50$, $-0.0096$ oracle-best). These sub-$0.02$ margins are read as parity rather than as the contribution: at this saturation, where the reference annotations carry errors that the anchor's own authors document and adjust their metric to tolerate~\cite{soucek2020transnetv2}, cross-domain F1 is governed by threshold calibration (Section~\ref{sec:res-threshold}). The parity is worth stating because of the condition under which it holds: \textsc{PERSIST} trains on ClipShots real transitions only, against an anchor trained on additional corpora with $85\%$ synthetic transitions. Table~\ref{tab:main} isolates that training-data factor directly, since a faithful single-rate TransNetV2 head on the same ClipShots-only budget (row~$\dagger$) trails the anchor on every modern cell at fixed-$0.50$ ($0.726$ in-domain), whereas \textsc{PERSIST} on that budget exceeds the reproduction on every modern dataset under the oracle-best protocol and on the diagnostic split, placing the anchor's edge in its training data rather than its architecture.

\begin{table*}[!tbp]
  \centering
  \footnotesize
  \setlength{\tabcolsep}{3.0pt}
  \renewcommand{\arraystretch}{1.12}
  \caption{\textbf{Tier~1 literature positioning:} F1 per dataset under the fixed-$0.50$ and oracle-best thresholds; best per column \textbf{bold}, second-best \underline{underlined} ($\dagger$ faithful single-rate TransNetV2 on our ClipShots-only data; $\ddagger$ three-seed probability-average ensemble).}
  \label{tab:main}
  \begin{tabular}{lcccccccc}
    \toprule
    Method & \multicolumn{2}{c}{ClipShots} & \multicolumn{2}{c}{BBC} & \multicolumn{2}{c}{AutoShot} & \multicolumn{2}{c}{HIST-GST12} \\
    \cmidrule(lr){2-3}\cmidrule(lr){4-5}\cmidrule(lr){6-7}\cmidrule(lr){8-9}
     & Fix-0.5 & Oracle & Fix-0.5 & Oracle & Fix-0.5 & Oracle & Fix-0.5 & Oracle \\
    \midrule
    Pixel difference~\cite{zabih1995feature}      & 0.0351 & 0.5602 & 0.0074 & 0.8316 & 0.0683 & 0.7252 & 0.0000 & 0.8631 \\
    HSV histogram~\cite{apostolidis2014fast}      & 0.5737 & 0.6019 & 0.7786 & 0.9244 & 0.6792 & 0.7765 & 0.1544 & 0.6550 \\
    Edge change ratio~\cite{zabih1995feature}     & 0.3510 & 0.3813 & 0.5552 & 0.5660 & 0.3505 & 0.6561 & 0.6019 & 0.7117 \\
    DeepSBD-AlexNet~\cite{hassanien2017large}     & 0.6843 & 0.7456 & 0.8202 & 0.8770 & 0.7539 & 0.7604 & 0.8535 & 0.8767 \\
    DeepSBD-ResNet18~\cite{hassanien2017large,tang2018fast} & 0.7555 & 0.7634 & 0.8622 & 0.8919 & 0.7303 & 0.7458 & 0.8611 & 0.8912 \\
    TNV2 (public)~\cite{soucek2020transnetv2}     & 0.7793 & 0.7794 & 0.9611 & \toptwo{0.9694} & \topone{0.7993} & \topone{0.8302} & 0.9204 & 0.9435 \\
    TNV2 (our repro)$^{\dagger}$   & 0.7260 & 0.7820 & 0.9517 & 0.9625 & \toptwo{0.7981} & 0.8029 & \topone{0.9305} & 0.9400 \\
    \midrule
    \textsc{PERSIST} (mean) & \toptwo{0.7913} & \toptwo{0.7941} & \toptwo{0.9661} & 0.9669 & 0.7878 & 0.8206 & 0.9227 & \toptwo{0.9458} \\
    \textsc{PERSIST} (multi-seed ens.)$^{\ddagger}$ & \topone{0.8023} & \topone{0.8075} & \topone{0.9700} & \topone{0.9707} & 0.7953 & \toptwo{0.8290} & \toptwo{0.9264} & \topone{0.9531} \\
    \bottomrule
  \end{tabular}
\end{table*}

\paragraph{Controlled comparison and attribution (Tier~2).}
\label{sec:res-tier2}
To remove training-protocol differences, Table~\ref{tab:controlled} compares the full model against the cue-only baseline of Section~\ref{sec:exp-tiers}, which shares \textsc{PERSIST}'s dual-rate backbone, optimiser, schedule, and seed and differs only in the head, replacing the persistence mechanism with the TransNetV2 descriptor-cue stack. This baseline therefore differs from the single-rate reproduction of Table~\ref{tab:main}: the latter controls training data with the original single-rate architecture, whereas this one controls the backbone to isolate the head.

Under the calibration-robust oracle-best protocol \textsc{PERSIST} leads the cue-only baseline on all four datasets (ClipShots $+0.012$, BBC $+0.006$, HISTORIAN-GST12 $+0.011$, AutoShot/SHOT $+0.007$). Under fixed-$0.50$ the lead is positive on ClipShots ($+0.065$) and BBC ($+0.018$), the large ClipShots gap partly reflecting the baseline's weaker calibration at $0.50$ rather than ranking quality, which is why the oracle-best column is the cleaner measure. AutoShot/SHOT and HISTORIAN-GST12 are the two fixed-$0.50$ exceptions, where the baseline's more conservative score scale yields advantages of $0.023$ and $0.005$ that both reverse under oracle-best.

The gain is not the discriminator's: removing it costs $0.001$ in-domain oracle-best F1 over the same three seeds (Section~\ref{sec:res-tier3}), well inside the $\pm0.003$ seed spread, so the ClipShots $+0.012$ is carried by the broader head design. The detector beats a TransNetV2-cue detector of the same backbone family under matched training, and the discriminator's distinctive value is the targeted pseudo-event suppression analysed next, not this benchmark increment.

\begin{table*}[!tbp]
  \centering
  \footnotesize
  \setlength{\tabcolsep}{4.0pt}
  \renewcommand{\arraystretch}{1.08}
  \caption{\textbf{Tier~2 controlled comparison:} the two rows share training data, optimiser, schedule, seeds, and evaluation pipeline and differ only in the head, with the best per column in bold and $\Delta$ the signed gain of \textsc{PERSIST} over the baseline. Entries are mean\,$\pm$\,population std over seeds $11$, $1$ and $21$.}
  \label{tab:controlled}
  \resizebox{\textwidth}{!}{%
  \begin{tabular}{lcccccccc}
    \toprule
     & \multicolumn{2}{c}{ClipShots} & \multicolumn{2}{c}{BBC} & \multicolumn{2}{c}{AutoShot} & \multicolumn{2}{c}{HIST-GST12} \\
    \cmidrule(lr){2-3}\cmidrule(lr){4-5}\cmidrule(lr){6-7}\cmidrule(lr){8-9}
    Method & Fix-0.5 & Oracle & Fix-0.5 & Oracle & Fix-0.5 & Oracle & Fix-0.5 & Oracle \\
    \midrule
    TNV2-cue baseline                & $.727{\pm}.006$ & $.782{\pm}.007$ & $.949{\pm}.003$ & $.961{\pm}.003$ & $\mathbf{.811}{\pm}.002$ & $.813{\pm}.003$ & $\mathbf{.928}{\pm}.004$ & $.935{\pm}.001$ \\
    \textsc{PERSIST}                 & $\mathbf{.791}{\pm}.004$ & $\mathbf{.794}{\pm}.003$ & $\mathbf{.966}{\pm}.002$ & $\mathbf{.967}{\pm}.001$ & $.788{\pm}.009$ & $\mathbf{.821}{\pm}.006$ & $.923{\pm}.007$ & $\mathbf{.946}{\pm}.003$ \\
    \midrule
    $\Delta$ (ours $-$ baseline)     & $+0.065$ & $+0.012$ & $+0.018$ & $+0.006$ & $-0.023$ & $+0.007$ & $-0.005$ & $+0.011$ \\
    \bottomrule
  \end{tabular}}
\end{table*}

\paragraph{Threshold behaviour and calibration.}
\label{sec:res-threshold}
Nor is the Tier-2 gain a capacity advantage: the cue baseline carries $18.18$M parameters against \textsc{PERSIST}'s $13.20$M ($0.09$M the persistence head). The fixed-$0.50$ and oracle-best protocols answer different questions, a deployment operating point versus a ranking upper bound, and should not be averaged. A full F1-versus-threshold sweep (supplementary material) shows the optimal threshold drifting for both models, confirming that cross-domain calibration is a property of the SBD setting rather than of either model. Expected calibration error follows the same pattern (full numbers in the supplementary material): an in-domain strength and a robustness gain over the matched cue reproduction, not a uniform advantage over the anchor.

\subsection{Component ablation}
\label{sec:res-tier3}

The ablation localises the contribution. Table~\ref{tab:ablation} reports the single-component ablation across all four datasets and both protocols, under the uniform val-best selection rule of Section~\ref{sec:exp-impl}; a fifth row removes all three latent-shaping components together.

Over three seeds the persistence discriminator's effect on aggregate F1 is mostly inside seed spread: removing it under-performs the full model on five of the eight cells, and only on AutoShot/SHOT at fixed-$0.50$ is the gap stable across all three paired seeds ($+0.025$). Its contribution shows instead where the mechanism is defined to act, confirmed by the per-subtype diagnostic (Section~\ref{sec:res-pertype}) and the latent-state analysis (Section~\ref{sec:res-latent}). Because this variant keeps the SIREN latent and a gate-free focal background-suppression loss and removes only the gates, it isolates what they add over the plain focal suppression a classifier could apply without them: at matched true-transition recall the same backbone trained this way fires on average $63$ more pseudo-event false positives, and is worse on two of three paired seeds (Section~\ref{sec:res-pertype}), since the targeted suppression term is defined through the gates and cannot be formed without them. The latent-shaping components, slow pathway, $\phi$ shaping, and dynamics auxiliary, are instead close to aggregate-F1-neutral, with small, not-consistently-signed per-cell differences (the $\phi$-shaping variant even edges the full model in-domain on ClipShots), and are validated in Sections~\ref{sec:res-pertype} and~\ref{sec:res-latent} rather than by the headline F1. That cell is the exception for them too: on AutoShot/SHOT at fixed-$0.50$ every ablation, the discriminator included, sits $0.019$ to $0.027$ below the full model, whereas on the other seven cells none moves it by more than $0.009$.

The fifth row confirms this at the configuration level: removing all three latent-shaping components at once stays within $0.003$ of the full model on every cell but that one, where it is $0.020$ lower.

\begin{table*}[!tbp]
  \centering
  \footnotesize
  \setlength{\tabcolsep}{4.0pt}
  \renewcommand{\arraystretch}{1.08}
  \caption{\textbf{Tier~3 single-component ablation} (uniform val-best selection), each row removing one component from the full model except the last which removes all three latent-shaping components together, with the best per column in bold. Entries are mean\,$\pm$\,population std over seeds $11$, $1$ and $21$; where arms tie at the printed precision all are bolded.}
  \label{tab:ablation}
  \resizebox{\textwidth}{!}{%
  \begin{tabular}{lcccccccc}
    \toprule
     & \multicolumn{2}{c}{ClipShots} & \multicolumn{2}{c}{BBC} & \multicolumn{2}{c}{AutoShot} & \multicolumn{2}{c}{HIST-GST12} \\
    \cmidrule(lr){2-3}\cmidrule(lr){4-5}\cmidrule(lr){6-7}\cmidrule(lr){8-9}
    Variant & Fix-0.5 & Oracle & Fix-0.5 & Oracle & Fix-0.5 & Oracle & Fix-0.5 & Oracle \\
    \midrule
    \textsc{PERSIST}                   & $.791{\pm}.004$ & $.794{\pm}.003$ & $.966{\pm}.002$ & $.967{\pm}.001$ & $\mathbf{.788}{\pm}.009$ & $\mathbf{.821}{\pm}.006$ & $\mathbf{.923}{\pm}.007$ & $.946{\pm}.003$ \\
    w/o discriminator                  & $.790{\pm}.005$ & $.793{\pm}.004$ & $.967{\pm}.001$ & $.968{\pm}.001$ & $.763{\pm}.020$ & $.816{\pm}.006$ & $.916{\pm}.012$ & $\mathbf{.949}{\pm}.004$ \\
    w/o $\phi$ shaping                 & $\mathbf{.794}{\pm}.002$ & $\mathbf{.796}{\pm}.002$ & $.966{\pm}.005$ & $.967{\pm}.005$ & $.769{\pm}.020$ & $.815{\pm}.003$ & $\mathbf{.923}{\pm}.012$ & $.947{\pm}.003$ \\
    w/o dynamics auxiliary             & $.787{\pm}.005$ & $.792{\pm}.000$ & $\mathbf{.969}{\pm}.000$ & $.969{\pm}.001$ & $.761{\pm}.030$ & $.812{\pm}.004$ & $\mathbf{.923}{\pm}.007$ & $.946{\pm}.003$ \\
    Single-rate (no slow path)         & $.788{\pm}.002$ & $.790{\pm}.004$ & $.966{\pm}.002$ & $.968{\pm}.002$ & $.765{\pm}.029$ & $.816{\pm}.005$ & $.919{\pm}.008$ & $\mathbf{.949}{\pm}.002$ \\
    w/o all three shaping components   & $.792{\pm}.003$ & $.795{\pm}.001$ & $\mathbf{.969}{\pm}.001$ & $\mathbf{.970}{\pm}.001$ & $.768{\pm}.030$ & $.818{\pm}.002$ & $.921{\pm}.009$ & $.947{\pm}.005$ \\
    \bottomrule
  \end{tabular}}
\end{table*}

\subsection{Pseudo-event suppression}
\label{sec:res-pertype}

Aggregate F1 averages a handful of pseudo-event frames into a sea of background and ordinary-transition frames, so the effect the mechanism targets becomes visible only once it is isolated. We isolate it on a $2{,}727$-video diagnostic split that renders each pseudo-event subtype and each transition type onto real held-out ClipShots shots, giving clips with genuine content, a known subtype label, and exact ground truth, none overlapping any detector's training data, and no detector (\textsc{PERSIST} included) trains on the rendered pseudo-events, except the hard-negative control below, so the split measures generalisation, not memorisation of synthetic negatives (construction in the supplementary material). These are the failure modes of Figure~\ref{fig:teaser} reproduced at scale. Every variant and the cue-only baseline see the identical clips, and the official TransNetV2 is scored separately at matched recall, since a single fixed threshold would place it at a different operating point.

Against the cue-only baseline the full system cuts false positives sharply: flash $329$ versus $508$, archival $270$ versus $404$, and text-overlay $80$ versus $399$ ($-35$, $-33$, and $-80\%$), at $0.50$, where the baseline is the more sensitive (recall $0.765$ against $0.726$). The advantage is not tied to one backbone configuration: at matched recall and a single seed, the persistence head fires fewer false positives than the cue head on both the single-rate ($775$ versus $812$) and the dual-rate backbone ($723$ versus $887$), the full model lowest of the four cells.

\begin{figure*}[!tb]
  \centering
  \includegraphics[width=0.82\linewidth]{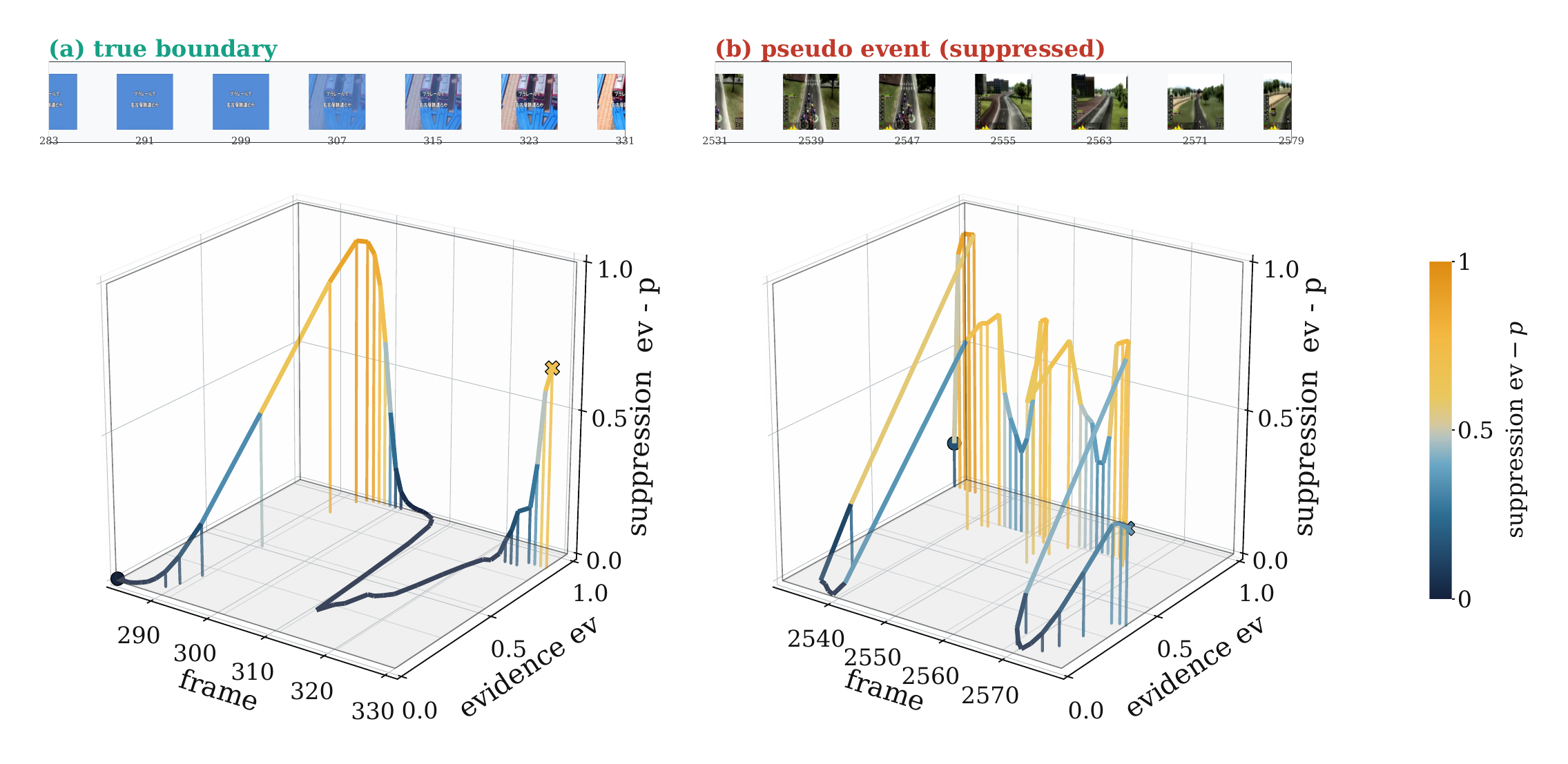}
  \caption{Space-time cube contrasting a true boundary with a pseudo event; the per-frame suppression $\mathrm{ev}-p$ is trajectory height (and, redundantly, lightness) above a no-suppression grey floor, with depth giving the local evidence $\mathrm{ev}$, so the pseudo-event trajectory rises high while the true boundary stays on the floor.}
  \label{fig:stc}
\end{figure*}

Geometrically, this selectivity reads as a space-time cube following the H\"agerstrand--Kraak convention~\cite{hagerstrand1970what,kraak2003space} (Figure~\ref{fig:stc}): a true boundary keeps $\mathrm{ev}$ and the final score together, so its trajectory stays on the no-suppression floor, whereas a pseudo event carries high evidence but a suppressed score, lifting its trajectory high and bright above the floor.

\begin{figure}[t]
  \centering
  \includegraphics[width=0.8\linewidth]{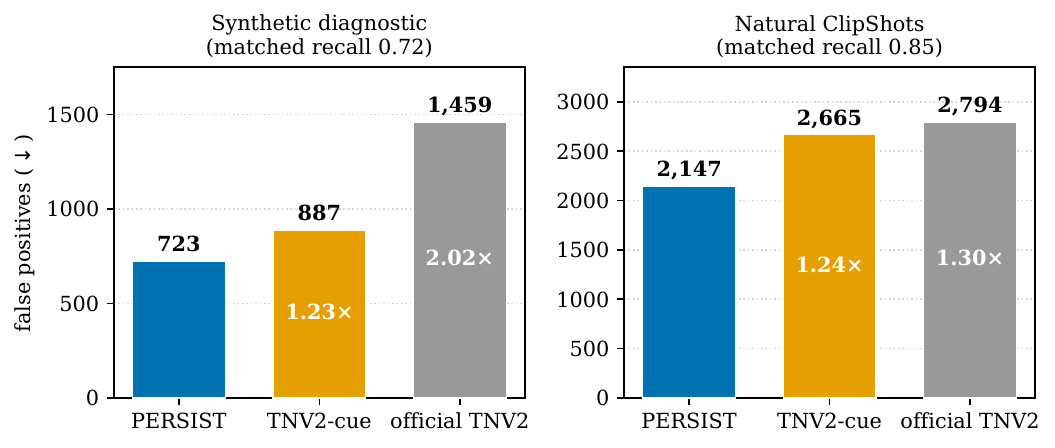}
  \caption{\textbf{False positives at matched true-transition recall:} with all three detectors tuned to the same true-transition recall, \textsc{PERSIST} fires the fewest on both splits; on the synthetic diagnostic the false positives are labelled pseudo-events, on natural ClipShots they are false detections on real non-boundary footage (ratios relative to \textsc{PERSIST} as $\times$; matched-recall counts, not the fixed-$0.50$ counts in the text).}
  \label{fig:payoff}
\end{figure}

The advantage survives the two checks that usually deflate such numbers. A raw count against the official TransNetV2 would be confounded by calibration, since at threshold $0.50$ the anchor sits at lower true-transition recall ($0.63$) than \textsc{PERSIST} ($0.73$) and fires fewer false positives merely by being less sensitive. Matching recall instead (the anchor at threshold $0.22$, recall $0.72$), it emits $1{,}459$ pseudo-event false positives to \textsc{PERSIST}'s $723$ and is worse on every subtype (flash $476$/$329$, text-overlay $420$/$80$, archival $456$/$270$, fast-pan $79$/$36$, scratch $28$/$8$). At equal sensitivity to genuine cuts, the reformulation cuts the anchor's pseudo-event count by about half. The same comparison on the real ClipShots test set gives \textsc{PERSIST} the fewest false positives of the three at matched recall $0.85$ ($2{,}147$ against $2{,}665$ and $2{,}794$); the precision gain is a property of real footage, not of the rendered split (Figure~\ref{fig:payoff}). Nor is the gain merely better negative supervision. Hard-negative fine-tuning of that cue baseline and the persistence test are two interventions on one starting point, both measured from it at matched recall. On the rendered diagnostic the hard negatives help more ($887\!\to\!603$ against $887\!\to\!723$), in the only detector fine-tuned on that split's own generator; on natural footage the ordering inverts, the formulation removing $518$ against $218$. Harder negatives buy accuracy on their own rendering; the reformulation buys what transfers.

The cost on ordinary transitions is narrow. On real-transition subtypes (supplementary material) the persistence test stays within $0.002$ F1 of the discriminator-free variant on clean cuts and is the strongest variant on dissolve. The two exceptions are jump cut and wipe: a jump cut preserves most scene content so the latent barely moves, and a wipe is a sharp sliding boundary a local cue tracks directly; both are genuine limitations of keying suppression on latent persistence, and both are better served by local cues.

\subsection{Latent-state analysis}
\label{sec:res-latent}

Two readouts, computed offline from saved per-frame latents on the ClipShots test set, make the persistence idea measurable; unlike the aggregate-F1 ablation, both are threshold-independent. The first concerns the latent itself. Averaged over $45{,}310$ boundary and $2.94$M non-boundary steps, the full model moves $\phi$ $1.49{\pm}0.13\times$ more at boundaries than elsewhere, above both shaping ablations ($1.31$ and $1.32$, three seeds) but inside seed spread. Retraining with one gate removed is sharper: at matched recall on natural footage, every gate removal raises false positives on all three paired seeds (return $+86$, transient $+135$, both at once $+101$). That is why these components are justified despite leaving almost no mark on aggregate F1 (Table~\ref{tab:ablation}).

The second concerns the decision. Grouped by ground-truth clip type, true cuts and model-flagged high-evidence pseudo-events arrive with almost the same raw evidence ($\mathrm{ev}\approx0.865$ versus $0.841$), yet the final score pulls them $2.7\times$ apart ($0.725$ versus $0.269$). The inference-time subtraction accounts for little of this ($\widetilde{\mathrm{ev}}=0.797$ versus $0.774$) and the joint gate fires almost identically on the two ($g_{\text{a}}{=}0.171$ versus $0.168$): the separation is carried by the base classifier that the shaping and the background-suppression loss have trained, with the gate serving as the interpretable read-out it is designed to be. Consistent with that design, forcing $g_{\text{a}}{=}0$ at inference changes the diagnostic pseudo-event count by only $7$ ($730$ versus $723$), whereas removing the gate-defined loss in training costs $48$ at the same seed and threshold, and $63$ at matched recall over three seeds: the gate's value is realised in training, and on the $3$M background frames the change gate is essentially zero.

\begin{figure*}[!tb]
  \centering
  \includegraphics[width=0.9\linewidth]{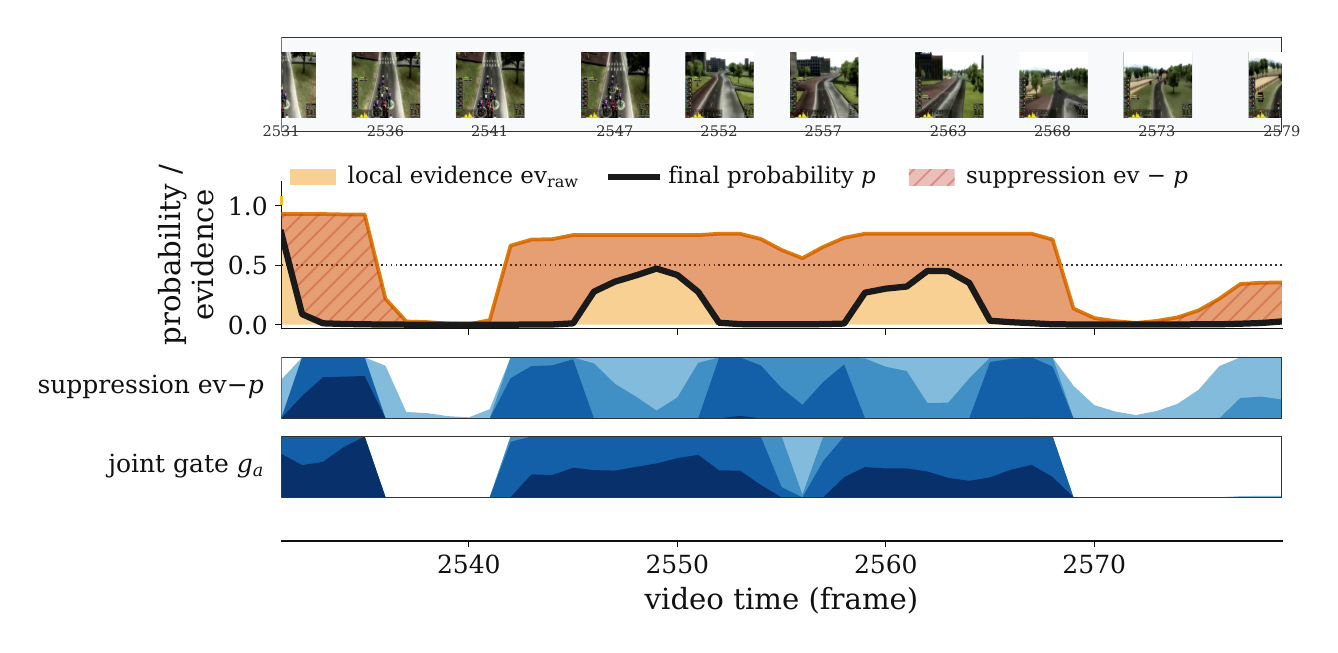}
  \caption{Per-frame decision atlas on a sustained pseudo event, where the local boundary evidence $\mathrm{ev}$ (filled; labelled $\mathrm{ev}_{\mathrm{raw}}$ in the panel) stays high while the final probability $p$ (solid) is held near zero to open the suppression band $\mathrm{ev}-p$, traced below by horizon rows for the suppression and the joint gate $g_a$ (darker meaning larger).}
  \label{fig:phi-cases}
\end{figure*}

Figure~\ref{fig:phi-cases} reads this out frame by frame on a sustained pseudo event: the local evidence stays high for several frames while the final probability is held near zero, opening a suppression band that darkens in lock-step with the discriminator's gate channel; on the true-cut control plotted the same way (supplementary material) the final probability rises with the evidence and stays high through the transition, and the boundary is kept.

\section{Conclusion}
\label{sec:conclusion}
\label{sec:summary}

We reformulated shot boundary detection as boundary semantic discrimination, a persistence test in the video's latent temporal state, instantiated as a dual-rate TransNetV2-style backbone, a FiLM-conditioned sinusoidal latent $\phi(t)$, and a three-gate persistence discriminator. The contribution rests on three independent layers of evidence: true cuts and high-evidence pseudo-events are indistinguishable by local evidence yet separated several-fold in the final score; \textsc{PERSIST} matches the strongest public detector from strictly less and weaker training data; and where the mechanism is decisive it suppresses a large, sign-consistent fraction of pseudo-events and makes the latent more responsive at boundaries. Where benchmark F1 has saturated, high-evidence pseudo-boundaries are where detectors still diverge, and suppressing them without losing recall is what \textsc{PERSIST} contributes.

\section*{Acknowledgements}
This research was funded in whole, or in part, by the Austrian Science Fund (FWF): [10.55776/DFH37]. For the purpose of Open Access, the author has applied a CC BY public copyright licence to any Author Accepted Manuscript (AAM) version arising from this submission.

\def\UrlBreaks{\do\/\do\-}  
\bibliography{egbib}

\clearpage
\runninghead{Lin et al.}{PERSIST: Supplementary Material}
\setcounter{section}{0}
\setcounter{figure}{0}
\setcounter{table}{0}
\setcounter{equation}{0}
\renewcommand{\thesection}{S\arabic{section}}
\renewcommand{\thefigure}{S\arabic{figure}}
\renewcommand{\thetable}{S\arabic{table}}
\renewcommand{\theequation}{S\arabic{equation}}
\renewcommand{\theHsection}{supp.\arabic{section}}
\renewcommand{\theHfigure}{supp.\arabic{figure}}
\renewcommand{\theHtable}{supp.\arabic{table}}
\renewcommand{\theHequation}{supp.\arabic{equation}}

\begin{center}
  {\LARGE\bfseries\sffamily\textcolor{bmv@sectioncolor}{Supplementary Material}\par}
  \vspace{0.35em}
  {\large\bfseries PERSIST: Persistent-State Discrimination for Shot Boundary Detection\par}
\end{center}
\vspace{0.75em}

\noindent This supplement is included in the same document as the main paper.
References of the form ``Table~\ref{supp:tab:datasets}'' or
``Figure~\ref{supp:fig:thresh}'' point within this supplement.

\section{Data, annotation, and the diagnostic split}
\label{supp:part-data}

This part documents what every reported number runs on, so the evaluation can be
reconstructed end to end: the scale of each benchmark, the film-editing
conventions behind the archival gradual-transition labels, and the construction
of the controlled pseudo-event diagnostic split that the main paper's
false-positive analysis depends on.

\subsection{Dataset statistics}
\label{supp:data}

Table~\ref{supp:tab:datasets} gives the full scale of every evaluation set.
The four benchmarks span almost two orders of magnitude in both video and
frame count, and the gradual-transition (GST) proportion varies from $0\%$
on the BBC fixed annotation to over $20\%$ on AutoShot/SHOT, so cross-domain
transfer is not dominated by a single transition regime. ClipShots~\cite{tang2018fast}
is collected from YouTube and Weibo (hard cases: hand-held vibration, large
object motion, occlusion); BBC Planet Earth~\cite{soucek2020transnetv2} is
eleven nature-documentary episodes under the widely adopted BBC annotation;
AutoShot/SHOT~\cite{zhu2023autoshot} is a short-video transfer set whose data
statistics differ from the other two; HISTORIAN-GST12 is a focused subset of
twelve HISTORIAN-derived archival videos~\cite{helm2022historian}, each with at
least one annotated gradual-transition span. All cross-domain evaluations use
a single ClipShots-trained model with no per-target fine-tuning.

\begin{table}[!htbp]
  \centering
  \footnotesize
  \setlength{\tabcolsep}{4.0pt}
  \renewcommand{\arraystretch}{1.2}
  \caption{\textbf{Evaluation datasets at a glance} (AST: abrupt shot transitions, cuts and jumps; GST: gradual transitions such as dissolve, fade, wipe, flip-over, swish pan).}
  \label{supp:tab:datasets}
  \begin{tabular}{lcrrrr}
    \toprule
    Dataset & Role & Videos & AST & GST & Frames \\
    \midrule
    ClipShots train~\cite{tang2018fast}  & training source & 4{,}876 & 110{,}575 & 34{,}810 & 26{,}649{,}843 \\
    \midrule
    ClipShots test~\cite{tang2018fast}   & in-domain test   & 500     & 4{,}903   & 2{,}301  & 3{,}241{,}806 \\
    BBC Planet Earth~\cite{soucek2020transnetv2} & transfer test & 11   & 4{,}708   & 136      & 809{,}827 \\
    AutoShot/SHOT~\cite{zhu2023autoshot} & transfer test   & 200     & 2{,}078   & 438      & 204{,}031 \\
    HISTORIAN-GST12~\cite{helm2022historian} & historical transfer & 12 & 1{,}377 & 191    & 196{,}931 \\
    \bottomrule
  \end{tabular}
\end{table}

\paragraph{Historical-annotation note.}
Because the original HISTORIAN annotations provide shot boundaries rather than
an explicit transition-type taxonomy, a clean cut/gradual split cannot be
derived from them directly. HISTORIAN-GST12 is therefore reported as a focused
subset that contains annotated gradual-transition spans, separately from any
broader historical annotation set; its role is to test transfer to
gradual-transition-bearing archival footage where damaged reels, scratches, and
splices make hand-crafted cues fragile. The reproducibility load of the main
comparison is carried by ClipShots, BBC Planet Earth, and AutoShot/SHOT, with
HISTORIAN-GST12 adding historical-domain evidence; the derived annotations and
conversion scripts are released with the paper.

\subsection{Gradual-transition annotation protocol (archival transfer set)}
\label{supp:gst-protocol}

The HISTORIAN-GST12 transfer set carries a transition-type annotation of its
gradual transitions, used here only for evaluation. The transition-type
annotation pass was carried out by an expert annotator at the Ludwig Boltzmann
Institute for Digital History, a co-author of this paper. We summarise its
conventions so that the gradual-transition labels are interpretable; the
definitions follow a standard film-editing reference~\cite{reisz1968technique}.

\paragraph{Shots and boundaries.}
A shot is defined in purely technical terms: a series of consecutive frames
played without interruption, delimited by an in-point and an out-point. A
shot boundary is the join where one out-point meets the next in-point, arising
either from in-camera editing (starting and stopping the camera) or from
post-production editing.

\paragraph{Transition-type categories.}
Each gradual transition is annotated as a separate shot segment (two shots are
typically visible during the transition), with the exception of fades (below).
The categories are:
\begin{itemize}\itemsep2pt
  \item \textbf{Dissolve}: one shot gradually fades into another through a temporary overlap of both images; bounded by the first and last frames in which both images are simultaneously visible.
  \item \textbf{Wipe}: a boundary line (vertical, horizontal, or diagonal) moves across the frame, one image ``pushing out'' the other without transparent overlap; in some stylistic variants this line is a soft, dissolve-like zone rather than a hard edge, but it is still labelled a wipe as long as the push-out characteristic holds.
  \item \textbf{Flip-over}: the frame appears to rotate or flip on an axis (like turning a page), distinguished from a wipe by a three-dimensional rotation effect rather than a flat sliding boundary.
  \item \textbf{Swish-pan}: an extremely rapid camera pan producing strong, intentional motion blur in which no distinct subject is recognisable, functioning as a transition; distinguished from a normal pan, which is a controlled movement tracking a subject or describing space.
  \item \textbf{Missing-frame}: one or more absent frames cause a visible ``jerk''; a small gap need not constitute a boundary, whereas a large gap with significant visual change does, a distinction that is partly subjective.
  \item \textbf{Fade-in / fade-out (exception)}: a fade is a transition between blank frames and a cinematic image, so the gradual brightening/darkening is not a separate segment but is annotated at the level of the shot as a whole; the adjacent blank frames belong to the shot. A fade-out followed immediately by a fade-in is split near the midpoint of the blank run into two shots.
\end{itemize}

\paragraph{Boundary frames and borderline cases.}
For a gradual transition the start and end frames are the first and last frames
in which the transition effect is visibly present. Borderline cases such as very
short dissolves or small missing-frame gaps must be judged visually and may
be read differently by different viewers; very short gradual transitions are
nonetheless labelled as such rather than collapsed to abrupt cuts, and
ambiguous cases are not flagged separately but assigned to the dissolve category
by visual interpretation, often additionally tagged with a superordinate frame
descriptor (\eg\ misexposure, blurred, or black/white/partial frame). This residual subjectivity at the boundary between a faint
gradual transition and a non-transition disturbance is exactly the regime in
which a local-change cue is unreliable, and it motivates the persistence test
of the main paper.

\subsection{Pseudo-event diagnostic split: construction}
\label{supp:synthdiag}

The diagnostic split used by the main paper's pseudo-event analysis is
not a separate corpus of mined events and not abstract noise: it
is built by programmatically rendering known transitions and known pseudo-events
onto real ClipShots shots, so that every clip has authentic photographic
content but a fully known ground truth. This gives the controlled per-subtype
labels that natural footage cannot, while keeping the underlying imagery real.

\paragraph{Source material and split hygiene.}
All $2{,}727$ diagnostic clips are rendered from shots drawn from the ClipShots
test videos. Every detector in this paper is trained on ClipShots
train only, so the diagnostic source material is held out from all
training; there is no train/diagnostic leakage for any model, including the
baselines. No detector other than the hard-negative control of
Section~\ref{supp:b4} is trained on labelled pseudo-events or on any rendered hard negative: the suppression \textsc{PERSIST}
exhibits here is learned purely from the persistence objective applied to real
ClipShots non-boundary frames, so the split measures generalisation of the
mechanism rather than memorisation of synthetic negatives.

\paragraph{Categories and how each is rendered.}
Each clip is assigned one category and rendered accordingly, with a per-sample
record storing its category, the source video and shot indices, and the
resulting ground-truth boundary interval. Two families are produced.
(i)~True transitions, used to verify recall: clean cuts (splicing two shots of the same
video) and jump cuts (eliding a temporal gap inside a single shot), and gradual transitions (dissolve, fade-in/out, wipe)
synthesised between shots. (ii)~Pseudo-events, used to measure false
positives, each a strong local discontinuity within a single continuous
shot so that the correct label is ``no boundary'': flash/illumination bursts,
fast camera pans, text/graphic overlays, archival degradation (grain, flicker,
brightness instability), and emulsion scratches. A subset of true transitions
is additionally corrupted with light hard-negative perturbations to mimic
real-world overlap of transition and damage.

\paragraph{Subtype counts in the test split.}
The split contains, by \texttt{synthetic\_type}: flash $510$, fast pan $453$,
text overlay $391$, archival $349$, scratch $182$ (the five pseudo-event
subtypes of the main paper's false-positive table); and clean cut $228$, jump
cut $181$, dissolve $130$, fade-out $123$, fade-in $116$, wipe $64$ (the
true-transition subtypes of Table~\ref{supp:tab:realf1}). Because the
comparison is read per subtype rather than as a single weighted aggregate, and
every detector is scored on the identical clips, the subtype mixture does not
bias the comparison between methods. The generator and manifests are
released with the paper.

\section{Method and training details}
\label{supp:part-method}

This part gives what is needed to reproduce the method and the comparison it is
read against: the rationale behind the three comparison tiers, and the complete
architecture, optimisation, and loss specification.

\subsection{Three-tier design: full rationale}
\label{supp:tiers}

The main paper summarises the three comparison tiers; we record the full
caveats here.

\paragraph{Tier~1 (literature positioning).}
All rows use the same evaluation script, target split, and $\pm2$-frame matching
tolerance, but training data, framework, and inference window differ across
rows. The official TransNetV2 model is trained on a larger, partly synthetic
mix (TRECVID IACC.3 with ClipShots; $\sim$$85\%$ synthetic transitions, of which
$35\%$ hard cuts and $50\%$ dissolves, and only $\sim$$15\%$ real ClipShots
transitions)~\cite{soucek2020transnetv2} and inferred with a $100$-frame window
at $27{\times}48$, whereas \textsc{PERSIST}
uses ClipShots real transitions only and a $T{=}64$ sliding window at the same
resolution. The training-side asymmetry runs against \textsc{PERSIST}, so parity
under a unified evaluator is a conservative reading. We mirror the disclaimer
of~\cite{soucek2020transnetv2} (``the presented results do not have to
correspond to the most optimal setting of related methods'') and the
framework caveat of~\cite{zhu2023autoshot}. The locally available RAI package
is not used as a main shot-level benchmark because its redistributed annotation
semantics are scene-oriented and are not directly comparable to the shot-level RAI
numbers reported in prior SBD papers. The released AutoShot model is excluded
from Tier~1 because its searched super-net is trained on the combined SHOT and
ClipShots sets~\cite{zhu2023autoshot}, so neither its architecture nor its
training data is comparable to a single fixed model trained on ClipShots alone. OmniShotCut~\cite{wang2026omnishotcut}
and TransVLM~\cite{chen2026transvlm} are excluded because they redefine the
output task (relational shot ranges and continuous transition segments) and
report on their own benchmarks, so a frame-level boundary-F1 comparison under
our evaluator would not be meaningful.

\paragraph{Tier~2 (controlled comparison).}
The cue-only baseline shares backbone, optimiser, schedule, seed, and absence of
synthetic augmentation with \textsc{PERSIST}, differing only in replacing the
persistence discriminator and $\phi$ shaping by the official TransNetV2
descriptor-cue stack (frame-similarity head, colour-histogram cue, many-hot
transition targets). It reaches ClipShots fixed-$0.50$ F1 $0.7182$ at seed $11$, a $0.061$
gap below the original paper's $0.779$ attributable to training-data and
framework differences; it is therefore a controlled alternative configuration
of the local pipeline, not a faithful reproduction of the public model.

\paragraph{Tier~3 (single-component ablation).}
All variants train on the same ClipShots split and share the backbone,
schedule, and seed, each removing exactly one component: (i)~without the
persistence discriminator ($\lambda_{\text{sup}}{=}0$, gated suppression
disabled at inference); (ii)~without $\phi$ shaping
($\lambda_{\phi}{=}\lambda_{\text{tv}}{=}0$, SIREN branch kept); (iii)~without
the dynamics auxiliary ($\lambda_{\text{dyn}}{=}0$); (iv)~single-rate backbone
(slow pathway and lateral fusion disabled). The fifth row in the main ablation
table removes all three latent-shaping components together.

\subsection{Implementation details}
\label{supp:impl}

Clips are resized to $27\times48$ and processed in $T{=}64$ sliding windows.
The backbone uses three stages with two cells per stage on each pathway, fast
base width $16$, slow base width $32$, slow temporal stride $8$, and lateral
channel ratio $0.25$. The SIREN body has $L{=}3$ FiLM-conditioned hidden
layers of width $128$ ($K{=}16$, $\omega_0{=}30$, conditioning dim $D_c{=}32$).
The persistence discriminator uses thresholds $\theta_{\text{c}}{=}0.6$,
$\theta_{\text{t}}{=}0.05$, $\theta_{\text{r}}{=}0.6$; temperatures
$\tau_{\text{c}}{=}0.05$, $\tau_{\text{t}}{=}0.03$, $\tau_{\text{r}}{=}0.1$;
offsets $\mathcal{D}{=}\{4,8,12\}$; and suppression strength $\alpha{=}0.4$.
Of these, only the return offsets $\mathcal{D}$ and the inference window length are
swept (Tables~\ref{supp:tab:horizon} and~\ref{supp:tab:winlen}); every other value above
is shared by all five test sets and is never tuned per dataset, which is what the
single-model protocol of the main paper requires.
The boundary-factor coefficient $\beta$ that blends the suppressed evidence into
the score is a single learned scalar, parameterised as $\beta{=}\sigma(\hat\beta)$
so it stays in $(0,1)$ and initialised to $0.15$. The final logit is recovered
from $p$ as $\ell_0$ and calibrated to $\ell{=}(\ell_0+b_{\text{cal}})/\tau_{\text{cal}}$,
with bias $b_{\text{cal}}$ a learned scalar initialised to $0$ and temperature
$\tau_{\text{cal}}{=}e^{\hat\tau}$ a learned positive scalar initialised to $1.0$
and clamped to $[0.7,4.0]$; $\beta$, $b_{\text{cal}}$, and $\tau_{\text{cal}}$ are
per-model scalars learned end-to-end.
Optimisation is SGD with momentum $0.9$ and weight decay $10^{-4}$. Loss
weights are $\lambda_{\text{b}}{=}1.0$, $\lambda_{\text{aux}}{=}1.0$,
$\lambda_{\text{siren}}{=}0.05$, $\lambda_{\phi}{=}0.2$,
$\lambda_{\text{tv}}{=}0.01$, $\lambda_{\text{dyn}}{=}0.01$ (curvature weight
$\mu{=}0.25$), $\lambda_{\text{bg}}{=}0.1$, $\gamma_{\text{bg}}{=}2$,
$\lambda_{\text{sup}}{=}0.05$; positive weights $w_{\text{cls}}{=}2$ (main),
$w_{\text{b}}{=}5$ (start/end), and $w_{\text{d}}{=}5$ (dynamics). The reported
model trains on the ClipShots train split only; no synthetic transitions are
mixed in, and all cross-domain evaluations are strict zero-shot transfers.
Training the full model takes roughly $34$ hours on four NVIDIA A40 GPUs (ten
epochs, effective batch $224$); inference keeps the standard TransNetV2 compute class,
since the SIREN latent and the gates add only a small per-frame head.

\paragraph{Training-loss terms.}
Let $y_t,y_t^{\text{s}},y_t^{\text{e}}\in\{0,1\}$ be the per-frame, start, and end
labels, $\ell(t)$ the calibrated final logit recovered from $p(t)$, and
$\mathrm{BCE}^{w}$ a binary cross-entropy with positive-class weight $w$. The three
groups of the total loss $\mathcal{L}=\mathcal{L}_{\text{cls}}+\mathcal{L}_{\phi}+\mathcal{L}_{\text{bg}}$
(main paper) are
\begin{align*}
  \mathcal{L}_{\text{cls}} &= \mathrm{BCE}^{w_{\text{cls}}}(\ell,y)
    + \lambda_{\text{aux}}\,\mathrm{BCE}^{w_{\text{cls}}}(\ell_{\text{base}},y)
    + \lambda_{\text{siren}}\,\mathrm{BCE}^{w_{\text{cls}}}(\ell_{\text{siren}},y) \\
    &\quad + \lambda_{\text{b}}\!\!\sum_{\bullet\in\{s,e\}}\!\!\mathrm{BCE}^{w_{\text{b}}}(\ell_{\bullet},y^{\bullet}),\\
  \mathcal{L}_{\phi} &= \lambda_{\phi}\,\|\phi-\mathbf{u}\|_2^2
    + \lambda_{\text{tv}}\,\overline{\|\Delta_x\phi\|_2^2}
    + \lambda_{\text{dyn}}\,\mathrm{BCE}^{w_{\text{d}}}(\ell_{\text{dyn}},y),
    \qquad \ell_{\text{dyn}}(t)=a\big(E(t)/\overline{E}-1\big)+b,\\
  \mathcal{L}_{\text{bg}} &= \lambda_{\text{bg}}\,\big\langle p^{1+\gamma_{\text{bg}}}\big\rangle_{\text{bg}}
    + \lambda_{\text{sup}}\,\big\langle \mathrm{ev}\,\mathrm{sg}(g_a\,g_c)\big\rangle_{\text{bg}},
\end{align*}
where $E(t)=\|\phi'(t)\|_2+\mu\|\phi''(t)\|_2$ is the analytic SIREN energy, $a,b$
are learned scale/bias, $\overline{\cdot}$ is the mean over $t$, and
$\langle\cdot\rangle_{\text{bg}}$ the mean over background frames ($y_t<0.5$), and
$\mathrm{sg}(\cdot)$ the stop-gradient operator, so the suppression term steers the
raw evidence rather than the gates (the extra change-gate factor concentrates it on
high-evidence frames); $\ell_{\text{siren}}$ and $\ell_{\text{dyn}}$
are training-only and never read at inference.

\section{Extended benchmark results and controls}
\label{supp:part-results}

This part shows the main-paper parity is robust rather than a tuning or confound
artefact: seed variation and ensembling, a matched single-rate reproduction, an
inference-window control, a persistence-horizon sensitivity check, an explicit
hard-negative baseline that isolates the source of the suppression, and the full
threshold-calibration behaviour.

\subsection{Multi-seed statistics and three-seed ensemble}
\label{supp:seeds}

Table~\ref{supp:tab:seeds} reports \textsc{PERSIST} over three training seeds
($11$, $1$, $21$; all val-best checkpoints, same protocol as the main paper).
Across all five test sets the per-seed standard deviation is at most $0.0095$
under fixed-$0.50$ and $0.0083$ under oracle-best, the latter on the synthetic
diagnostic; over the four benchmarks of the main paper's Table~1 it is at most
$0.0064$ (ClipShots $0.0031$, BBC $0.0013$ oracle-best). The controlled comparison and
the ablation of the main paper report mean\,$\pm$\,population std over the same three
seeds, with the per-seed values in Table~\ref{supp:tab:seed3}. A three-seed ensemble
(per-frame probability average) improves every dataset over the seed mean and,
beyond the single-model comparison, reaches or exceeds the official
TransNetV2 anchor on four of the five test sets and is within $0.002$ on the fifth, closing the AutoShot oracle-best gap
($0.829$ vs $0.830$) and exceeding the anchor on the synthetic diagnostic
($0.631$ vs $0.557$). The ensemble uses three models and is therefore reported
separately, not as the headline single-model result.

\begin{table}[!htbp]
  \centering
  \footnotesize
  \setlength{\tabcolsep}{3.5pt}
  \renewcommand{\arraystretch}{1.15}
  \caption{\textbf{Three-seed mean\,$\pm$\,std and ensemble} (fix-0.5 / oracle-best F1; ENS is the three-seed probability-average ensemble; official TransNetV2 oracle-best shown for reference).}
  \label{supp:tab:seeds}
  \begin{tabular}{lccc}
    \toprule
    Dataset & mean$\pm$std (fix / oracle-best) & ENS (fix / oracle-best) & TNV2 (oracle-best) \\
    \midrule
    ClipShots      & $0.7913{\pm}.004$ / $0.7941{\pm}.003$ & $0.8023$ / $0.8075$ & $0.7794$ \\
    BBC            & $0.9661{\pm}.002$ / $0.9669{\pm}.001$ & $0.9700$ / $0.9707$ & $0.9694$ \\
    AutoShot/SHOT  & $0.7878{\pm}.009$ / $0.8206{\pm}.006$ & $0.7953$ / $0.8290$ & $0.8302$ \\
    HIST-GST12     & $0.9227{\pm}.007$ / $0.9458{\pm}.003$ & $0.9264$ / $0.9531$ & $0.9435$ \\
    synth diag     & $0.5193{\pm}.002$ / $0.5924{\pm}.008$ & $0.5484$ / $0.6311$ & $0.5568$ \\
    \bottomrule
  \end{tabular}
\end{table}

\paragraph{Robustness to the aggregation rule.}
The ensemble's standing does not depend on how the three seeds are combined. Probability averaging, logit averaging (a standard log-opinion pool), and the geometric mean of the per-frame scores agree to within $0.003$ oracle-best F1 on every benchmark (for example ClipShots $0.8075$/$0.8068$/$0.8075$ and AutoShot $0.8290$/$0.8312$/$0.8319$ for the three rules). On AutoShot, the single cell where probability averaging sits just below the anchor at oracle-best ($0.829$ vs $0.830$), the log-opinion-pool variants land marginally above it ($0.831$ and $0.832$); these differences of at most $0.002$ are within the per-seed standard deviation and we read them, like the other cross-domain margins, as parity rather than as improvements, and under the fixed-$0.50$ protocol AutoShot stays below the anchor for every combiner. We therefore report the natural probability average and draw no leaderboard claim from the ensemble.

\subsection{Single-stream TransNetV2 reproduction}
\label{supp:b3}

The ``TNV2 (our repro, single-rate)'' row of the main Tier-1 table is a
faithful TransNetV2 classifier head (descriptor flatten map, frame-similarity
and colour-histogram cues, many-hot targets, transition-weighted loss) on a
single-rate TransNetV2-cell backbone with the slow pathway and lateral fusion
removed, trained on ClipShots real transitions only under our budget (no
synthetic rendering). It thus matches the official architecture family while
controlling the training data to ours. Oracle F1: ClipShots $0.782$, BBC
$0.963$, AutoShot/SHOT $0.803$, HIST-GST12 $0.940$, synthetic diagnostic
$0.558$. It trails the official public model on every modern cell at fixed-$0.50$, the gap
attributable to the official model's $85\%$-synthetic training corpus rather
than its architecture, and is exceeded by \textsc{PERSIST} under the oracle-best protocol on all five test
sets under the same training budget, isolating the contribution of the
persistence mechanism from both architecture and training data.

\paragraph{Window-length control (64 vs 100 frames).}
The official TransNetV2 runs $100$-frame inference windows whereas we use $64$.
To rule out a window-length confound, we retrain and evaluate both
\textsc{PERSIST} and the single-rate reproduction at a $100$-frame window
(Table~\ref{supp:tab:winlen}). At $100$ frames neither model improves over its
$64$-frame counterpart: across all four datasets the oracle-best difference is at
most $0.010$ F1, and the longer window is in fact slightly worse at the
fixed threshold on AutoShot and HISTORIAN-GST12. Matching the anchor's window
therefore does not help us, so the $64$-frame results reported throughout are
not advantaged by the window choice, and the \textsc{PERSIST}-versus-TransNetV2
comparison is not confounded by it.

\begin{table}[!htbp]
  \centering
  \footnotesize
  \setlength{\tabcolsep}{4.0pt}
  \renewcommand{\arraystretch}{1.15}
  \caption{\textbf{Inference window-length control} (single-model, seed $11$, fix-0.5 / oracle-best F1).
  Neither model benefits from the $100$-frame window the official anchor uses.}
  \label{supp:tab:winlen}
  \begin{tabular}{llcccc}
    \toprule
    Model & Window & ClipShots & BBC & AutoShot & HIST-GST12 \\
    \midrule
    \textsc{PERSIST}             & $64$  & 0.7937/0.7964 & 0.9672/0.9678 & 0.7992/0.8280 & 0.9319/0.9494 \\
    \textsc{PERSIST}             & $100$ & 0.7890/0.7938 & 0.9725/0.9729 & 0.7785/0.8182 & 0.9171/0.9419 \\
    \midrule
    TNV2 repro (single-rate)     & $64$  & 0.7260/0.7820 & 0.9517/0.9625 & 0.7981/0.8029 & 0.9305/0.9400 \\
    TNV2 repro (single-rate)     & $100$ & 0.7287/0.7845 & 0.9558/0.9648 & 0.8016/0.8016 & 0.9266/0.9331 \\
    \bottomrule
  \end{tabular}
\end{table}

\subsection{Persistence-horizon sensitivity}
\label{supp:horizon}

The return gate reads the latent at a set of multi-scale offsets $\mathcal{D}$
around each candidate; the main paper uses $\mathcal{D}=\{4,8,12\}$ frames, and a
natural question is whether the suppression depends on this horizon. We retrain
\textsc{PERSIST} from scratch with the offsets halved to $\{2,4,6\}$ and doubled
to $\{8,16,24\}$, changing nothing else, and evaluate ClipShots F1 and the
synthetic pseudo-event false-positive count (Table~\ref{supp:tab:horizon}). Both
are nearly flat: ClipShots oracle-best F1 moves by at most $0.0035$ and the total
pseudo-event count by at most $46$ out of $723$, while every horizon stays far
below the cue baseline ($1{,}391$) and the official TransNetV2 at matched recall
($1{,}459$). The default offsets are marginally best on both axes, and the only
subtype that shifts appreciably is flash. The multi-scale aggregation therefore
removes the need to tune a per-dataset offset, and the suppression is a property
of the persistence test rather than of one specific horizon.

\begin{table}[!htbp]
  \centering
  \footnotesize
  \setlength{\tabcolsep}{6pt}
  \renewcommand{\arraystretch}{1.15}
  \caption{\textbf{Persistence-horizon sensitivity} (single-model, seed $11$): ClipShots fix-0.5 / oracle-best F1 and the total synthetic pseudo-event false positives ($\downarrow$, threshold $0.5$). The default offsets are marginally best and all three stay far below the cue baseline ($1{,}391$) and the matched-recall TransNetV2 ($1{,}459$).}
  \label{supp:tab:horizon}
  \begin{tabular}{lcc}
    \toprule
    Return-gate offsets $\mathcal{D}$ & ClipShots (fix / oracle-best) & synth FP \\
    \midrule
    $\{2,4,6\}$ (short)    & 0.7849 / 0.7929 & 769 \\
    $\{4,8,12\}$ (default) & 0.7937 / 0.7964 & 723 \\
    $\{8,16,24\}$ (long)   & 0.7890 / 0.7949 & 766 \\
    \bottomrule
  \end{tabular}
\end{table}

\subsection{Hard-negative supervision versus the persistence formulation}
\label{supp:b4}

A natural question is whether \textsc{PERSIST}'s pseudo-event suppression comes from the
persistence formulation or simply from better negative supervision: a detector with the
same dual-rate context could instead be trained on pseudo-event hard negatives.
We test this directly. Starting from the cue-only baseline (the dual-rate
TransNetV2-cue head of the controlled comparison, with no SIREN latent and no persistence
discriminator), we finetune it on ClipShots real transitions mixed with rendered
pseudo-event hard negatives for two epochs; we call it the hard-negative baseline. \textsc{PERSIST}
itself is trained on no hard negatives at all.

\begin{table}[!htbp]
  \centering
  \footnotesize
  \setlength{\tabcolsep}{4.5pt}
  \renewcommand{\arraystretch}{1.15}
  \caption{\textbf{Negative supervision vs the persistence formulation.} Synthetic-diagnostic
  and natural false positives ($\downarrow$; both at matched true-transition recall,
  $0.7257$ on the diagnostic and $0.8499$ on the ClipShots test set), with cross-domain oracle-best F1. The hard-negative baseline is the
  cue head trained with rendered pseudo-event hard negatives; \textsc{PERSIST} uses none. All rows are seed $11$.}
  \label{supp:tab:b4}
  \resizebox{\columnwidth}{!}{%
  \begin{tabular}{lcccc}
    \toprule
    Model & ClipShots oracle & synth FP (flash/overlay) & natural FP & xdom oracle (BBC/AS/HIST) \\
    \midrule
    \textsc{PERSIST}                       & 0.7964 & 723 (329/80)   & \textbf{2147} & 0.968/0.828/0.949 \\
    Cue + hard negatives                   & 0.7868 & \textbf{603} (146/196)           & 2447          & 0.966/0.808/0.936 \\
    Cue baseline (no hard neg.)            & 0.7733 & 887 (396/181)          & 2665          & 0.963/0.810/0.936 \\
    \bottomrule
  \end{tabular}}
\end{table}

Table~\ref{supp:tab:b4} reports the comparison. Explicit hard-negative training does help, and on the rendered diagnostic it helps
more than the persistence formulation does. Read at matched recall from the shared
starting point, the hard negatives take the cue baseline from $887$ to $603$ while the
persistence formulation takes it to $723$. That ordering is expected: the hard-negative
baseline trained on that exact rendering distribution, whereas \textsc{PERSIST} saw no
rendered pseudo-event at all. On natural footage, where neither has that advantage, the
ordering inverts: from the same baseline the formulation removes $518$ false positives
($2{,}665\!\to\!2{,}147$) against the hard negatives' $218$ ($2{,}665\!\to\!2{,}447$).
The reduction is also uneven across subtypes: at matched recall the hard-negative baseline
beats \textsc{PERSIST} on flash ($146$ vs $329$) and archival degradation ($207$ vs $270$)
but stays far worse on text overlay ($196$ vs $80$) despite $4{,}024$ text-overlay hard
negatives in its training pool, and is even worse there than its own un-finetuned parent
($181$). That is the signature of in-distribution fitting rather than the broadly
transferable suppression \textsc{PERSIST} provides without any hard negatives. What harder
negatives buy is accuracy on the rendering they were cut from; what the reformulation buys
is the part that transfers.

\subsection{Threshold-calibration sweep}
\label{supp:thresh}

\begin{figure*}[!tbp]
  \centering
  \includegraphics[width=\linewidth]{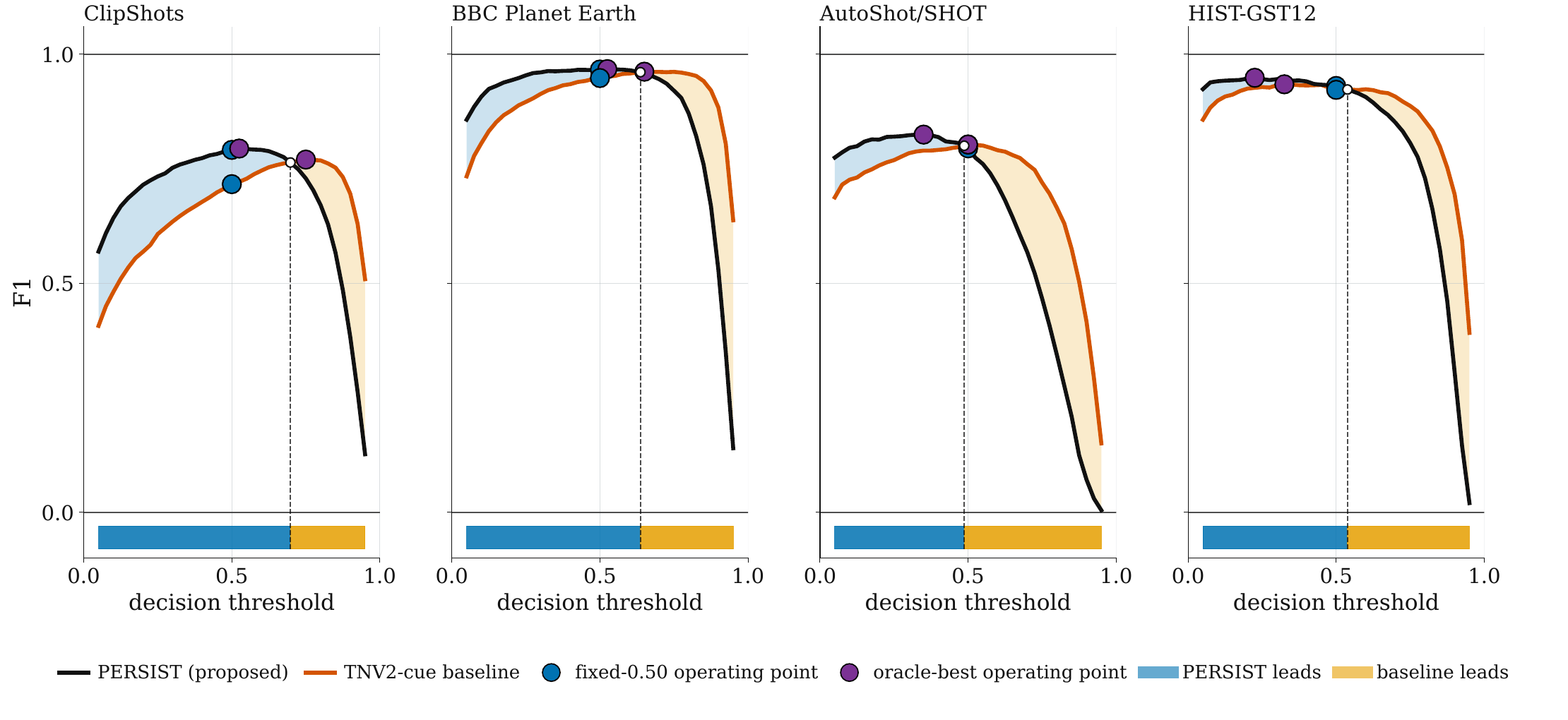}
  \caption{\textbf{F1 versus decision threshold} on the four datasets (\textsc{PERSIST}: black curve; within-protocol TNV2-cue baseline: orange curve), with markers for the fixed-$0.50$ and per-dataset oracle-best operating points and a strip beneath each axis shaded blue where \textsc{PERSIST} leads and yellow where the baseline leads.}
  \label{supp:fig:thresh}
\end{figure*}

Figure~\ref{supp:fig:thresh} shows the full F1-versus-threshold curve for
\textsc{PERSIST} and the cue-only baseline on each dataset, with the
fixed-$0.50$ and oracle-best operating points marked. The optimal threshold
drifts across datasets for both models (\textsc{PERSIST}: $0.54$ on ClipShots,
$0.26$ on HISTORIAN-GST12; baseline: $0.72$ and $0.33$), confirming that
cross-domain threshold calibration is a property of the SBD setting rather than
of either model. Of the two markers on each curve, the fixed-$0.50$ point is the
deployable operating threshold and the oracle-best point the per-split ranking
upper bound.

On expected calibration error (ECE; ten equal-width probability bins over all
aligned frames, positives being frames inside annotated transition spans) the
picture is specific. \textsc{PERSIST} has the
best in-domain value (ClipShots $0.0088$, against $0.0112$ for the official
TransNetV2 and $0.0209$ for our cue reproduction), and is far better calibrated
than the cue reproduction across the modern benchmarks (BBC $0.019$ vs $0.029$,
AutoShot $0.030$ vs $0.054$). Against the official anchor the cross-domain
comparison is mixed: \textsc{PERSIST} is better on the historical set ($0.025$
vs $0.031$) but behind on BBC and AutoShot ($0.019/0.030$ vs $0.015/0.022$).
Calibration is therefore an in-domain strength and a robustness gain over the
matched reproduction, not a uniform advantage over the anchor.

\section{Mechanism analysis}
\label{supp:part-mechanism}

This part substantiates the interpretability claims with the per-variant evidence
behind the main paper's mechanism discussion: real-transition recall is not
sacrificed, the false-positive reduction decomposes cleanly by component, and the
latent state and per-frame decision behave as the persistence account predicts.
It closes with the leave-one-gate-out study, which isolates what each gate
contributes, and with the limitations of the study.

\subsection{Per-seed values behind the controlled comparison and the ablation}
\label{supp:seedtab}

The controlled comparison and the ablation of the main paper report the three-seed
mean\,$\pm$\,population std. Table~\ref{supp:tab:seed3} gives the individual seed values behind them, so that the
paired-seed statements of the main paper can be checked directly and the spread of each arm can be
read per corpus rather than through a summary. Every arm was trained at seeds $11$, $1$ and $21$ and
evaluated under the same val-best rule.

\begin{table*}[!htbp]
  \centering
  \footnotesize
  \setlength{\tabcolsep}{5pt}
  \renewcommand{\arraystretch}{1.12}
  \caption{\textbf{Per-seed F1} behind the three-seed means of Tables~2 and~3 of the main paper,
  given as seed $11$\,/\,seed $1$\,/\,seed $21$ under each protocol. AutoShot/SHOT at fixed-$0.50$ carries
  by far the largest seed spread of the four corpora, which is why the statements that touch it are
  qualified in the main paper.}
  \label{supp:tab:seed3}
  \resizebox{\textwidth}{!}{%
  \begin{tabular}{lcccc}
    \toprule
    Variant & ClipShots & BBC & AutoShot/SHOT & HIST-GST12 \\
    \midrule
    \multicolumn{5}{l}{\emph{fixed-$0.50$}} \\
    \textsc{PERSIST}                   & $.794/.785/.795$ & $.967/.964/.967$ & $.799/.776/.788$ & $.932/.916/.921$ \\
    w/o discriminator                  & $.784/.790/.797$ & $.966/.967/.968$ & $.790/.743/.756$ & $.928/.919/.900$ \\
    w/o $\phi$ shaping                 & $.792/.791/.797$ & $.970/.959/.969$ & $.795/.764/.748$ & $.936/.907/.926$ \\
    w/o dynamics auxiliary             & $.790/.792/.779$ & $.969/.969/.969$ & $.784/.719/.781$ & $.926/.913/.931$ \\
    Single-rate (no slow path)         & $.790/.785/.790$ & $.968/.966/.964$ & $.801/.730/.764$ & $.928/.920/.908$ \\
    w/o all three shaping components   & $.790/.795/.790$ & $.969/.967/.970$ & $.791/.725/.788$ & $.932/.910/.921$ \\
    TNV2-cue baseline                  & $.718/.733/.729$ & $.948/.945/.952$ & $.810/.813/.810$ & $.926/.923/.934$ \\
    \midrule
    \multicolumn{5}{l}{\emph{oracle-best}} \\
    \textsc{PERSIST}                   & $.796/.790/.796$ & $.968/.965/.968$ & $.828/.812/.821$ & $.949/.941/.947$ \\
    w/o discriminator                  & $.791/.791/.799$ & $.967/.967/.970$ & $.812/.811/.824$ & $.952/.950/.944$ \\
    w/o $\phi$ shaping                 & $.799/.793/.797$ & $.971/.959/.970$ & $.818/.814/.812$ & $.949/.944/.949$ \\
    w/o dynamics auxiliary             & $.792/.792/.791$ & $.969/.969/.970$ & $.816/.807/.814$ & $.948/.942/.948$ \\
    Single-rate (no slow path)         & $.793/.785/.793$ & $.970/.968/.965$ & $.822/.811/.815$ & $.951/.946/.949$ \\
    w/o all three shaping components   & $.794/.797/.794$ & $.969/.969/.971$ & $.819/.815/.821$ & $.952/.941/.947$ \\
    TNV2-cue baseline                  & $.773/.790/.782$ & $.963/.957/.963$ & $.810/.816/.814$ & $.936/.934/.935$ \\
    \bottomrule
  \end{tabular}}
\end{table*}

\subsection{Per-subtype real-transition F1}
\label{supp:realtrans}

Table~\ref{supp:tab:realf1} reports F1 per real-transition subtype on the
$2{,}727$-video synthetic diagnostic split, for the full model, all five
ablation variants, and the cue-only baseline. It is the per-variant detail
behind the summary in the main paper, which already establishes that the
persistence test does not penalise genuine abrupt cuts and that only jump cut
and wipe favour the frame-similarity cue; we do not repeat those figures here.
The added ablation rows make one further point, visible only when this table is
read against the false-positive counts in the main paper: the single-rate
variant slightly raises real-transition F1 on several subtypes (clean
cut, jump cut, fade-in, wipe) while simultaneously raising the
hard-negative false-positive counts, i.e.\ dropping the slow pathway simply
makes the detector fire more readily.

\begin{table*}[!tbp]
  \centering
  \footnotesize
  \setlength{\tabcolsep}{4.0pt}
  \renewcommand{\arraystretch}{1.15}
  \caption{\textbf{Real-transition F1 ($\uparrow$) per synthetic subtype} (per column, best in \textbf{bold}, second-best \underline{underlined}). The w/o $\phi$ shaping row is evaluated at the val-best checkpoint of Section~4.1 of the main paper; three rows are tied for second on fade-in.}
  \label{supp:tab:realf1}
  \resizebox{\textwidth}{!}{%
  \begin{tabular}{lcccccc}
    \toprule
    Method & clean cut (228) & jump cut (181) & dissolve (130) & fade-out (123) & fade-in (116) & wipe (64) \\
    \midrule
    \textsc{PERSIST}                 & 0.901          & 0.123          & \topone{0.880} & 0.926          & \toptwo{0.903} & 0.581 \\
    w/o discriminator                & 0.903          & 0.118          & 0.866          & 0.922          & \toptwo{0.903} & 0.619 \\
    w/o $\phi$ shaping               & 0.905   & \toptwo{0.132} & 0.870  & 0.926  & \toptwo{0.903} & 0.583 \\
    w/o dynamics auxiliary           & 0.904          & 0.111          & 0.860          & \topone{0.941} & 0.900          & 0.559 \\
    Single-rate (no slow path)       & \toptwo{0.906} & 0.127  & 0.851          & 0.930          & \topone{0.908} & \toptwo{0.634} \\
    w/o all three shaping components & \topone{0.911} & 0.115          & \toptwo{0.871} & \toptwo{0.937} & 0.900          & 0.500 \\
    TNV2-cue baseline                & 0.897          & \topone{0.172} & 0.834          & 0.910          & 0.856          & \topone{0.678} \\
    \bottomrule
  \end{tabular}}
\end{table*}

\subsection{Per-component pseudo-event decomposition}
\label{supp:fpdecomp}

The main paper reports the headline pseudo-event reductions and the
mechanism\,$\times$\,slow-pathway contrast; here we record the per-component
decomposition of the total false-positive count behind them. Across the five
hard-negative subtypes the total pseudo-event count is $723$ for the full
model, and it moves cleanly by component: removing the persistence
discriminator adds $48$ ($723\!\to\!771$) and removing the slow pathway adds
$117$ ($723\!\to\!840$), so both genuinely suppress pseudo-events; removing the
dynamics auxiliary adds $15$ and removing $\phi$ shaping adds
$18$ ($723\!\to\!741$). These counts are at threshold $0.50$ and seed $11$. Over three seeds and at
matched true-transition recall, which removes the calibration confound, both latent-shaping
ablations cost suppression on every paired seed: removing the dynamics auxiliary adds $68$
($794.3{\pm}40.6$ against the full model's $726.3{\pm}6.2$) and removing all three shaping
components adds $79$ ($805.0{\pm}34.2$), three of three paired seeds in each case. The latent-shaping components and the discriminator
therefore play distinct, not interchangeable, roles. Table~\ref{supp:tab:synth}
gives the full per-variant, per-subtype false-positive counts behind this
decomposition and behind the headline reductions quoted in the main paper.

\begin{table}[!tbp]
  \centering
  \footnotesize
  \setlength{\tabcolsep}{4.0pt}
  \renewcommand{\arraystretch}{1.15}
  \caption{\textbf{Pseudo-event false-positive counts ($\downarrow$) on the synthetic diagnostic split}, per hard-negative subtype with sample sizes in the header; among the model variants the best is in \textbf{bold} and the second-best \underline{underlined}. All rows are seed $11$ at threshold $0.50$.}
  \label{supp:tab:synth}
  \begin{tabular}{lccccc}
    \toprule
    Method & flash (510) & fast pan (453) & overlay (391) & archival (349) & scratch (182) \\
    \midrule
    \textsc{PERSIST}             & \topone{329} & 36          & 80          & \toptwo{270} & \toptwo{8} \\
    w/o discriminator                   & 363          & 37          & 82          & 282          & \topone{7} \\
    w/o $\phi$ shaping                  & \toptwo{339} & \toptwo{34} & 68   & 292  & \toptwo{8} \\
    w/o dynamics auxiliary              & 374          & \topone{33} & \topone{64} & \topone{259} & \toptwo{8} \\
    Single-rate (no slow path)          & 409          & 40          & 82          & 301          & \toptwo{8} \\
    w/o all three shaping components    & 384          & 37          & \toptwo{66} & 288          & 11 \\
    \midrule
    TNV2-cue baseline                   & 508          & 58          & 399         & 404          & 22 \\
    \bottomrule
  \end{tabular}
\end{table}

\subsection{Latent-state diagnostics: $\phi$ continuity and gate firing}
\label{supp:latent}

These are the full tables behind the latent-state analysis of the main paper.
Table~\ref{supp:tab:phicont} reports the ratio of the mean one-step latent change at boundary
versus non-boundary steps; over three seeds the full model reaches
$1.49{\pm}0.13$ against $1.31{\pm}0.16$ without the discriminator and $1.32{\pm}0.14$ without
$\phi$ shaping, an ordering that follows the design but by margins inside seed
spread. Table~\ref{supp:tab:gatefire}
groups every ClipShots test frame by ground-truth clip type, with pseudo-events
restricted to the frames the model itself scores as high-evidence: true cuts and
high-evidence pseudo-events carry near-identical raw evidence $\mathrm{ev}$ yet the
final score $p$ separates them $2.7\times$, the joint gate fires almost identically
on the two, and the change gate is essentially zero on background, so the criterion
acts through the trained classifier rather than as a selective inference-time filter.
The two diagnostics count different units. The $\phi$-continuity readout
(Table~\ref{supp:tab:phicont}) averages the one-step displacement
$\|\phi(t{+}1){-}\phi(t)\|$, so each clip window contributes $T{-}1$ steps and a
$\pm2$-frame ring around every annotated span is held out as a separate
neighbourhood bucket, whereas Table~\ref{supp:tab:gatefire} tallies frames.
The $45{,}310$ boundary steps quoted in the main paper are therefore slightly
fewer than the $48.3$k AST-plus-GST transition frames here ($9.7$k${+}38.6$k):
the difference is this step-versus-frame convention, not a different boundary
tolerance.

\begin{table}[!htbp]
  \centering
  \footnotesize
  \setlength{\tabcolsep}{5pt}
  \renewcommand{\arraystretch}{1.15}
  \caption{\textbf{$\phi(t)$ continuity diagnostic (ClipShots), three seeds:} ratio of the mean one-step latent change at boundary steps to non-boundary steps, per seed and as mean\,$\pm$\,population std over seeds $11$/$1$/$21$, all at the val-best checkpoint. The absolute displacement is not comparable across runs, since nothing fixes the scale of $\phi$; only the within-run ratio is. The full model is highest, but the margins over both shaping ablations lie inside seed spread, so this readout corroborates the design rather than establishing it.}
  \label{supp:tab:phicont}
  \begin{tabular}{lcccc}
    \toprule
    Model & seed $11$ & seed $1$ & seed $21$ & mean\,$\pm$\,std \\
    \midrule
    \textsc{PERSIST}   & $1.41$ & $1.38$ & $1.67$ & $\mathbf{1.49}{\pm}0.13$ \\
    w/o discriminator   & $1.09$ & $1.44$ & $1.39$ & $1.31{\pm}0.16$ \\
    w/o $\phi$ shaping  & $1.49$ & $1.14$ & $1.33$ & $1.32{\pm}0.14$ \\
    \bottomrule
\end{tabular}
\end{table}

\begin{table}[!htbp]
  \centering
  \footnotesize
  \setlength{\tabcolsep}{4pt}
  \renewcommand{\arraystretch}{1.15}
  \caption{\textbf{Gate firing by ground-truth clip type} (full model, ClipShots): mean gate activations, raw and suppressed evidence, and final score per category.}
  \label{supp:tab:gatefire}
  \begin{tabular}{lccccc}
    \toprule
    Category (frames) & $g_{\text{c}}$ & $g_{\text{a}}$ & $\mathrm{ev}$ & $\widetilde{\mathrm{ev}}$ & $p$ \\
    \midrule
    background ($3.0$M)        & $0.000$ & $0.000$ & $0.013$ & $0.013$ & $0.005$ \\
    true cut / AST ($9.7$k)    & $0.912$ & $0.171$ & $0.865$ & $0.797$ & $0.725$ \\
    gradual / GST ($38.6$k)    & $0.595$ & $0.112$ & $0.621$ & $0.576$ & $0.605$ \\
    pseudo-event ($119$k)      & $0.896$ & $0.168$ & $0.841$ & $0.774$ & $0.269$ \\
    \bottomrule
  \end{tabular}
\end{table}

\subsection{Leave-one-gate-out: what each gate contributes}
\label{supp:loo}

The three gates enter the discriminator as a product, so no single gate can be
isolated by reading its activation alone. We therefore retrain the model with one
gate removed from the product, that is, held at $1$ so that it constrains neither
the background-suppression loss nor the inference-time adjustment, and leave
everything else unchanged. Three arms are trained: without
the transient-impulse gate, without the return gate, and with the change gate
alone. Each arm is run at seeds $11$, $1$, and $21$, and every arm is scored at
the same matched true-transition recall as the full model, so the comparison is
free of the calibration confound that affects fixed-threshold counts.

\begin{table}[!htbp]
  \centering
  \footnotesize
  \setlength{\tabcolsep}{4pt}
  \renewcommand{\arraystretch}{1.15}
  \caption{\textbf{Leave-one-gate-out, false positives at matched
  true-transition recall} (three seeds, mean\,$\pm$\,population std; lower is better).
  ``Paired'' counts the seeds on which the arm is worse than the full model trained
  with the same seed. On natural footage every gate is load-bearing on every paired
  seed; on the rendered diagnostic the return gate is, while the other two arms are
  worse on two of three seeds. The main paper quotes the seed-$11$ member of the full-model
  row, $723$ and $2{,}147$, where this table reports the mean over the three seeds.}
  \label{supp:tab:loo}
  \begin{tabular}{lcccc}
    \toprule
    & \multicolumn{2}{c}{synthetic diagnostic} & \multicolumn{2}{c}{natural ClipShots} \\
    \cmidrule(lr){2-3}\cmidrule(lr){4-5}
    Arm & FP & paired & FP & paired \\
    \midrule
    \textsc{PERSIST} (all three gates) & $\mathbf{726.3}{\pm}6.2$   & --    & $\mathbf{2207.0}{\pm}118.9$ & --    \\
    w/o transient-impulse gate         & $780.3{\pm}69.7$           & $2/3$ & $2342.0{\pm}112.8$          & $3/3$ \\
    w/o return gate                    & $779.7{\pm}32.3$           & $3/3$ & $2293.3{\pm}64.9$           & $3/3$ \\
    change gate only                   & $781.0{\pm}56.8$           & $2/3$ & $2308.0{\pm}101.6$          & $3/3$ \\
    \bottomrule
  \end{tabular}
\end{table}

\begin{table}[!htbp]
  \centering
  \footnotesize
  \setlength{\tabcolsep}{3.5pt}
  \renewcommand{\arraystretch}{1.15}
  \caption{\textbf{Leave-one-gate-out per pseudo-event subtype} (synthetic diagnostic, matched
  true-transition recall, mean\,$\pm$\,population std over seeds $11$/$1$/$21$; lower is better).
  The fourth row removes the discriminator entirely and is the endpoint of the series; the last
  row is the identically trained cue baseline, included so the recall-matched per-subtype
  comparison of Section~4.4 of the main paper can be read here. Signs are not uniform across
  subtypes: every leave-one-out arm costs flash and archival suppression, while text overlay
  improves when a gate is dropped. Against the cue baseline the aggregate margin is $-160.7$
  (paired on all three seeds) and holds on flash, fast pan, text overlay and scratch, but
  archival degradation is level ($234.7$ against $234.0$).}
  \label{supp:tab:loosub}
  \begin{tabular}{lcccccc}
    \toprule
    Arm & flash & fast pan & overlay & archival & scratch & total \\
    \midrule
    \textsc{PERSIST}            & $\mathbf{341.3}{\pm}13.4$ & $\mathbf{32.3}{\pm}2.6$ & $110.0{\pm}22.2$ & $\mathbf{234.7}{\pm}27.1$ & $\mathbf{8.0}{\pm}0.8$ & $\mathbf{726.3}{\pm}6.2$ \\
    w/o transient gate          & $398.0{\pm}7.0$  & $35.7{\pm}3.4$ & $\mathbf{88.7}{\pm}18.9$ & $246.0{\pm}54.3$ & $12.0{\pm}2.2$ & $780.3{\pm}69.7$ \\
    w/o return gate             & $371.3{\pm}23.7$ & $32.0{\pm}1.6$ & $105.3{\pm}31.5$ & $262.7{\pm}52.7$ & $8.3{\pm}1.2$ & $779.7{\pm}32.3$ \\
    change gate only            & $361.7{\pm}4.6$  & $36.7{\pm}3.1$ & $97.7{\pm}4.6$   & $276.7{\pm}54.0$ & $8.3{\pm}2.6$ & $781.0{\pm}56.8$ \\
    w/o discriminator           & $378.3{\pm}49.4$ & $38.3{\pm}0.9$ & $110.0{\pm}22.1$ & $252.7{\pm}26.4$ & $10.0{\pm}2.4$ & $789.3{\pm}57.4$ \\
    \midrule
    cue baseline                & $435.7{\pm}56.1$ & $38.3{\pm}2.1$ & $166.7{\pm}10.5$ & $234.0{\pm}40.6$ & $12.3{\pm}0.5$ & $887.0{\pm}88.2$ \\
    \bottomrule
  \end{tabular}
\end{table}

Tables~\ref{supp:tab:loo} and~\ref{supp:tab:loosub} report the totals and the per-subtype breakdown. Two things follow. First, the gates are not redundant: removing them costs suppression, and on natural footage it does so on every paired seed, which is
the evidence the main paper cites in its Section~4.5. Second, the cost is not additive: on natural
footage, dropping the transient-impulse gate alone is worse than dropping both it
and the return gate, and on the rendered diagnostic the three arms fall within one
standard deviation of one another. The gates therefore interact rather than
contributing independent increments, which is what the product form of Eq.~(8) of
the main paper implies: the criterion tests a conjunction, so removing one factor
changes what the remaining factors have to learn.

\subsection{Decision atlas: true-cut control}
\label{supp:decision-atlas}

\begin{figure*}[!tbp]
  \centering
  \includegraphics[width=\linewidth]{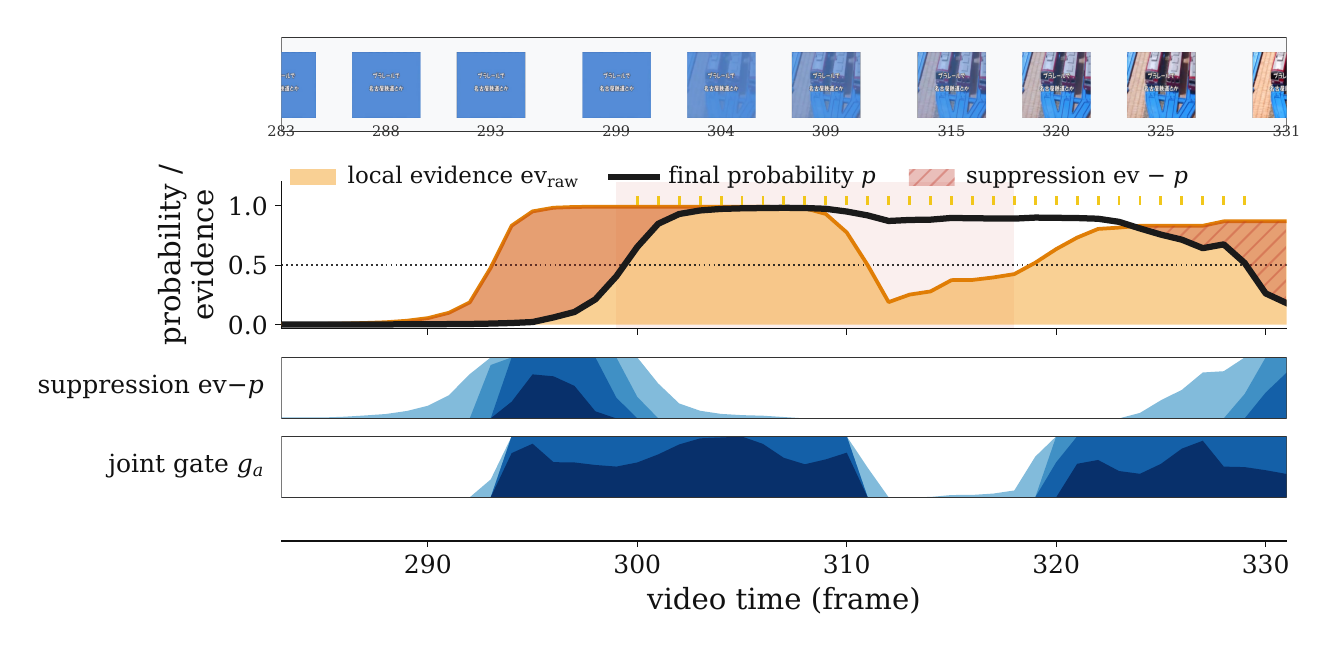}
  \caption{\textbf{Decision atlas, true-cut control} in the same per-frame signals and style as the main-paper figure: the final probability rises with the evidence and stays high through the transition, in contrast to the pseudo event shown in the main paper, where the score is held near zero.}
  \label{supp:fig:atlas}
\end{figure*}

Figure~\ref{supp:fig:atlas} is the true-cut control for the main paper's
decision atlas: on a genuine boundary the final probability rises with the
evidence and stays high through the transition, so the boundary is kept. The
gate read-outs fire around the boundary much as they do on the pseudo event,
consistent with the gate-firing analysis of the main paper; the two cases are
separated by the final score, which is held near zero on the pseudo event and
stays high here.

\subsection{Limitations}
\label{supp:limitations}

Five limitations bound what this study establishes. First, the persistence test is
keyed on latent movement, so it is weakest exactly where a true boundary moves the latent
little or where a local cue already suffices: jump cuts preserve most scene content and
wipes are sharp sliding boundaries, and the main paper reports both as the two
real-transition subtypes on which the mechanism does not help. Second, training uses a
single corpus, ClipShots real transitions only; the cross-domain results are therefore
transfer measurements, not evidence that the formulation is corpus-independent, and a
multi-corpus study is left open. Third, the temporal model is a dual-rate convolutional
backbone; state-space families such as S4 or Mamba would be a natural substitute for the
slow pathway and are not evaluated here, so no claim is made that the dual-rate design is
the best carrier of the persistence test. Fourth, efficiency is reported as batched
throughput ($21.7$k frames/s against the official anchor's $28.8$k, batch $8$, $64$- and
$100$-frame windows respectively, on one A100). Fifth, the correspondence between the
criterion and its implementation is partial in two ways: the post-transition settling
condition of Eq.~(1) of the main paper is not given a read-out of its own, the transient and return
gates testing instead whether the latent change fails to hold across the candidate; and at
inference the return gate is satisfied at almost every frame, its multi-scale similarity
having a tenth percentile above $0.98$ in every category. The conjunction's selectivity is
therefore realised through the training objective rather than through per-frame filtering,
which is what Section~4.5 of the main paper measures and what Figures~\ref{supp:fig:system}
and~\ref{supp:fig:system-b} show: between a true boundary and a within-shot frame the change
read-out moves from $1.00$ to $0.00$ while the transient and return read-outs barely move.

\section{Interactive visual-analytics tool}
\label{supp:system}

\begin{figure*}[!t]
  \centering
  \includegraphics[width=\linewidth]{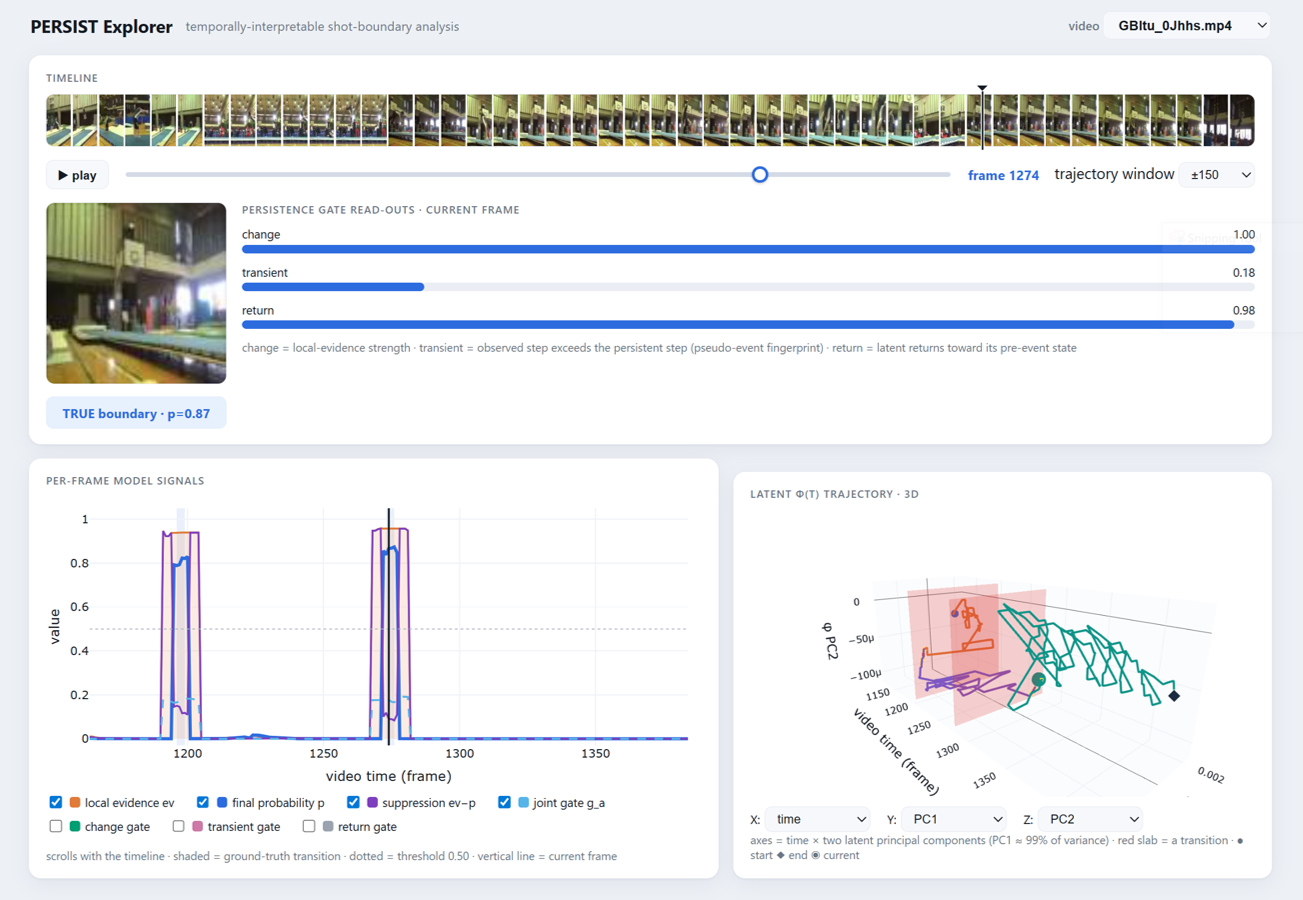}
  \caption{\textbf{\textsc{PERSIST~Explorer}, an interactive visual-analytics tool} built on the method's per-frame outputs, scrubbed here to a true shot boundary: the linked filmstrip timeline, per-frame signal panel, rotatable 3D latent-$\phi(t)$ trajectory, and frame inspector with the three gate read-outs all update together as the analyst moves through time.}
  \label{supp:fig:system}
\end{figure*}

\begin{figure*}[!t]
  \centering
  \includegraphics[width=\linewidth]{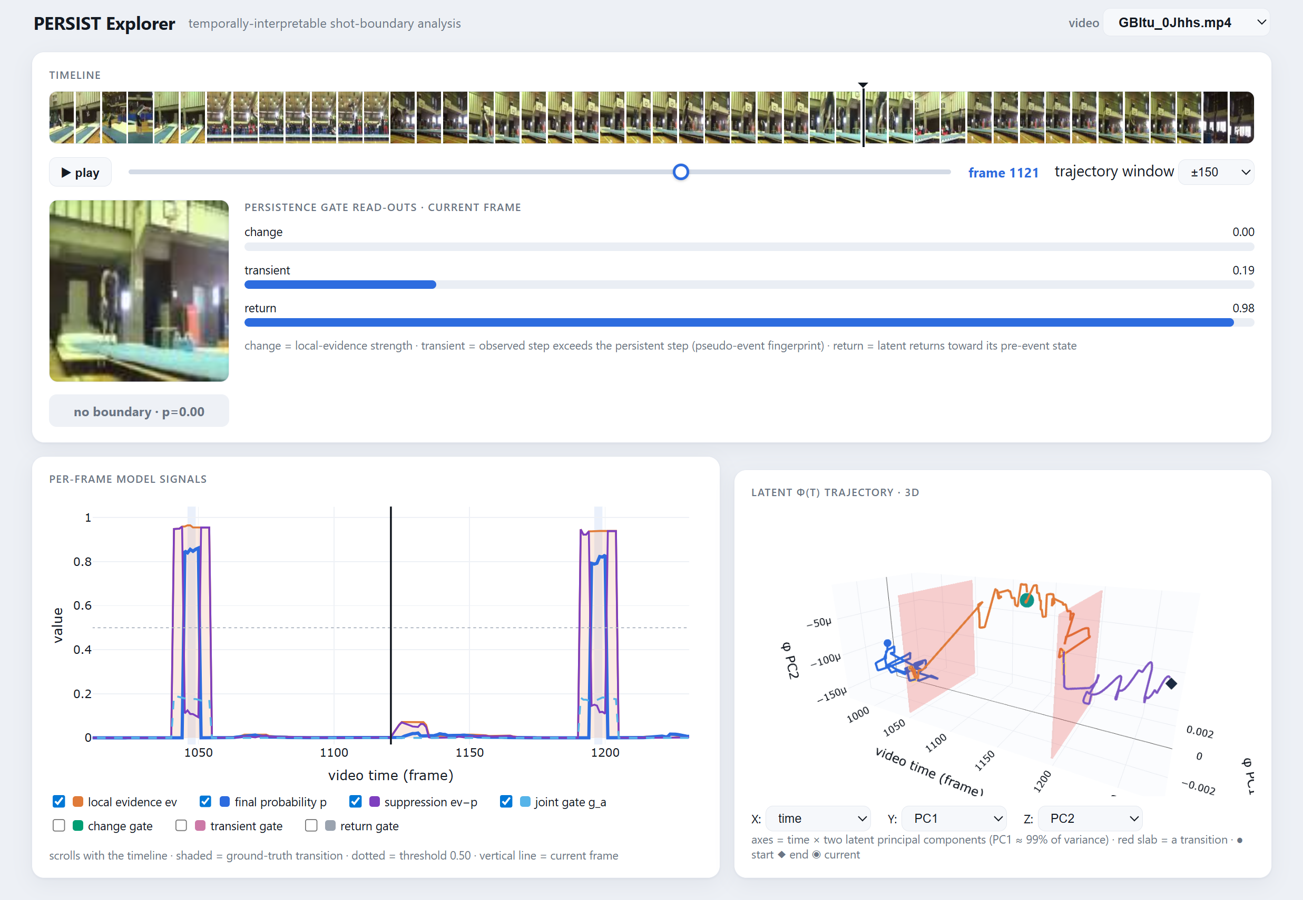}
  \caption{\textbf{The same \textsc{PERSIST~Explorer} session at a non-boundary frame.} Scrubbed to a frame inside a shot, the change gate reads zero, the suppression band stays closed, and the latent-$\phi(t)$ trajectory rests within its within-shot cluster, the quiescent counterpart to the boundary frame of Figure~\ref{supp:fig:system}.}
  \label{supp:fig:system-b}
\end{figure*}

The per-frame quantities the method exposes (local evidence, final probability,
suppression, the gate read-outs, and the continuous latent $\phi(t)$) are the
building blocks of a temporally interpretable view of a video. We build on them
\textsc{PERSIST~Explorer}, an interactive tool (Figures~\ref{supp:fig:system}
and~\ref{supp:fig:system-b}) in
which an analyst scrubs a timeline and linked views update together: a filmstrip,
the per-frame signal panel, a rotatable 3D latent-$\phi(t)$ trajectory, and a
frame inspector with the three gate read-outs and the boundary-versus-pseudo-event
verdict. In the latent view the trajectory rests in a stable cluster within each
shot and jumps at transitions; the space-time cube of the main paper is one
such view. It targets analysts such as archive and film scholars who need to see
why a frame is marked, runs in a browser on saved per-frame outputs, and is
released with the paper.

\end{document}